%% file: neurips_2024.tex
\documentclass{article}

\usepackage[final]{neurips_2024}

\usepackage[utf8]{inputenc} 
\usepackage[T1]{fontenc}    
\usepackage[
    colorlinks,
    linkcolor={red!80!black},
    citecolor={blue!50!black},
    urlcolor={blue!80!black}]{hyperref}       
\usepackage{url}            
\usepackage{booktabs}       
\usepackage{amsfonts}       
\usepackage{nicefrac}       
\usepackage{microtype}      
\usepackage{xcolor}         
\usepackage{csquotes}

\usepackage{graphicx,wrapfig} 
\usepackage{caption}
\usepackage[binary-units]{siunitx} 
\usepackage{tabularx} 
\usepackage{makecell} 
\usepackage{graphicx} 
\usepackage{todonotes}
\usepackage[subtle]{savetrees}
\usepackage{tikz}
\usetikzlibrary{shapes}
\usetikzlibrary{patterns}
\usetikzlibrary{positioning}
\usepackage[outline]{contour}

\title{Navigating Extremes:\\ Dynamic Sparsity in Large Output Spaces}

\author{%
  Nasib Ullah\textsuperscript{1} \quad
  Erik Schultheis\textsuperscript{1} \quad
  Mike Lasby\textsuperscript{2} \quad
  Yani Ioannou\textsuperscript{2} \quad
  Rohit Babbar\textsuperscript{1,3} \\
  \\
  \textsuperscript{1}Department of Computer Science, Aalto University, Helsinki, Finland \\
  \texttt{\{nasibullah.nasibullah, erik.schultheis, rohit.babbar\}@aalto.fi} \\
  \textsuperscript{2}Schulich School of Engineering, University of Calgary, Calgary, AB, Canada \\
  \texttt{\{mklasby, yani.ioannou\}@ucalgary.ca} \\
  \textsuperscript{3}Department of Computer Science, University of Bath, Bath, UK \\
  \texttt{rb2608@bath.ac.uk}
}

\begin{document}

\maketitle

\begin{abstract}
In recent years, Dynamic Sparse Training (DST) has emerged as an alternative to post-training pruning 
for generating efficient models. In principle, DST allows for a more memory efficient training process,
as it maintains sparsity throughout the entire training run. However, current DST implementations fail to capitalize on this in practice. Because sparse matrix multiplication is much less efficient than dense matrix multiplication on GPUs, most
implementations simulate sparsity by masking weights. 
In this paper, we leverage recent advances in semi-structured sparse training to apply DST in the domain of classification
with large output spaces, where memory-efficiency is paramount. With a label space of possibly millions of candidates,
the classification layer alone will consume several gigabytes of memory. Switching from a dense to a fixed fan-in 
sparse layer updated with sparse evolutionary training (SET); however, severely hampers training convergence, especially
at the largest label spaces. We find that poor gradient flow from the sparse classifier to the dense text encoder make it difficult to learn good input representations.
By employing an intermediate layer or adding an auxiliary training objective, we recover most of the generalisation performance of the dense model. 
Overall, we demonstrate the applicability and practical benefits of DST in a challenging domain --- characterized by a highly skewed label distribution 
that differs substantially from typical DST benchmark datasets --- which enables end-to-end training with millions of labels on commodity hardware.

\end{abstract}


\section{Introduction}

Recent research \cite{frankle2018lottery,frankle2020linear,malach2020proving,chen2021lottery} has demonstrated that densely-connected neural networks contain sparse subnetworks --- often dubbed \textquote{winning lottery tickets} --- that can deliver performance comparable to the full networks but with substantially reduced compute and memory demands. Unlike conventional techniques that start with a trained dense model and employ iterative pruning or one-shot pruning, Dynamic Sparse Training (DST) \cite{mocanu2018scalable,evci2020rigging,liu2021we} initializes a sparse architecture and dynamically explores subnetwork configurations through periodic pruning and regrowth, typically informed with heuristic \textit{saliency} criteria such as weight and gradient magnitudes. This approach is particularly advantageous in scenarios constrained by a fixed memory budget during the training phase, making DST viable across various domains \cite{chen2021lottery,dietrich2021towards,liu2021efficient}. For instance, in reinforcement learning \cite{tan2022rlx2,graesser2022state}, DST has been shown to significantly outperform traditional dense models. Additionally, models trained using DST often exhibit enhanced robustness \cite{guo2018sparse,ozdenizci2021training,chen2021sparsity,diffenderfer2021winning,liu2022deep}. However, the application of DST comes with challenges, notably prolonged training times; for example, RigL \cite{evci2020rigging} and ITOP \cite{liu2021we} require up to five and two times as many optimization steps during training, respectively, to match the generalisation performance of dense networks at high sparsity levels ($\geq80\%$). The prolonged training time in these works is often linked to the need for \emph{in-time overparameterization} \cite{liu2021we} and poor gradient flow in sparse networks. Recent advances \cite{evci2022gradient,price2021dense,curci2021truly,rajeshkumar2024progressive} aimed at improving gradient flow have been introduced to mitigate these extended training durations, enhancing the practicality of DST methodologies.

In this paper, we investigate the integration of DST into \emph{extreme multi-label classification} (XMC) \cite{Bhatia16,babbar2017dismec,prabhu2018parabel}.
XMC problems are characterized by a very large label space,
with hundreds of thousands to millions of labels, often in the same order of magnitude as the number of training
examples. 
The large label space in such problems makes calculating logits for every label a very costly operation. Consequently, 
 contemporary XMC methodologies \cite{jiang2021lightxml,you2019attentionxml,kharbanda2022cascadexml,dahiya2021siamesexml,dahiya2023ngame,zhang2021fast, schultheis2022speeding, qaraei2024meta} utilize modular and sampling-based techniques to achieve sublinear compute costs.
However, these strategies do not help in addressing the immense memory requirement associated with the classification layer,
which can be enormous: for an embedding dimension of 768, one million labels lead to a memory consumption of about 12 GB
taking into account weights, gradients, and optimizer state.\footnote{Using mixed precision with \texttt{torch.amp} has little
benefit here, because optimizer states and model parameters are still maintained in 32 bit, and only down-converted to speed-up
matrix multiplications.}
Memory efficiency in XMC has been pursued in the context of sparse \emph{linear} models~\citep{babbar2017dismec,babbar2019data,yen2016pd}
or by using label-hashing~\citep{medini2019extreme}, but such methods do not yield predictive performance competitive with
modern transformer-based deep networks. \citet{schultheis2023towards} demonstrated that applying a DST method
to the extreme classification layer can lead to substantial memory savings at marginal accuracy drops; however, that work
presupposed the existence of fixed, well-trained document embeddings which output the hidden representations used by the classifier, whereas in a realistic
setting these need to be trained jointly.

Recently, \citet{jain2023renee} demonstrated that full end-to-end training of
XMC models can be very successful, given sufficient computational resources. To
make this accessible to consumer-grade hardware, we propose to switch the dense
classification layer to a DST-trained sparse layer. Not only does this result
in a training procedure that allows XMC models to be trained in a GPU-memory
constrained setting, but it also provides an evaluation of DST algorithms
outside typical, well-behaved benchmarks. This is particularly important since
recent works \citep{liu2023sparsity,jaiswal2024compressing} have found that sparse
training algorithms that appear promising on standard benchmark datasets may
fail to produce adequate results on actual real-world tasks. As such, we introduce
XMC problems --- with their long-tailed label distribution~\citep{jain2016extreme,buvanesh2023enhancing, schultheis2024generalized}, 
missing labels~\citep{jain2016extreme,qaraei2021convex,schultheis2022missing, schultheis2021unbiased}, and
general training data scarcity issues~\citep{babbar2019data} --- as a new setting
to challenge current sparsity approaches.


\begin{figure}[bt]
  \centering
  \input{figures/architectures}\hskip-0.5ex%
  \scalebox{0.4}{\input{figures/intro_plot.pgf}}
  \caption{Model configurations and performance comparisons at various sparsity levels. The left panel illustrates our model configurations: `S' represents a semi-structured fixed fan-in sparse layer, `W' denotes an intermediate layer, and `Aux' refers to an auxiliary head of meta-classifiers. These configurations help maintain performance as the label space size increases from 31K to 670K and beyond. The right panel demonstrates the comparative precision at 1 for our model against other methods across increasing levels of sparsity on the Amazon670K dataset.}
  \label{fig:intro-figure}
\end{figure}
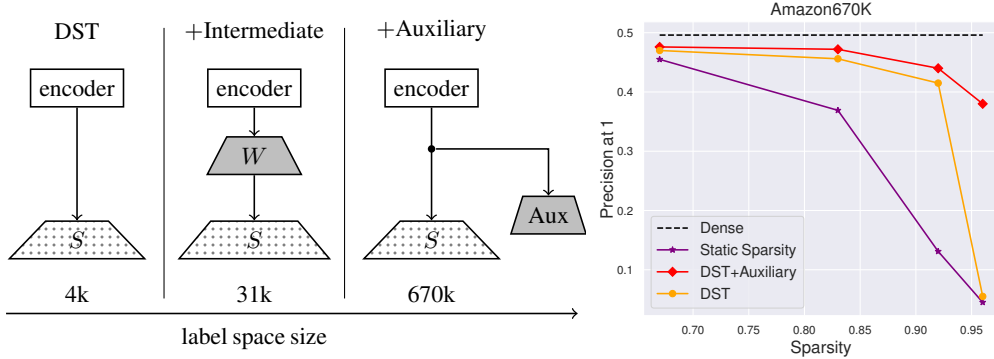

In fact, direct application of existing DST methods yields unsatisfactory results on XMC tasks due to typically noisy data and poor gradient signal propagation through the sparse classifier, slowing training convergence to an extent that it is not practically useful.
Consequently, we follow \citet{schultheis2023towards} and adapt the model architecture
by integrating an intermediate layer that is larger than the embedding
from the encoder but still significantly smaller than
the final output layer. While this was found to be sufficient to achieve good 
results with fixed encodings,  
if the encoder is a trainable transformer~\citep{devlin2018bert,sanh2019distilbert}
for label spaces with more than one hundred thousand elements, particularly at high levels of sparsity. 
The primary challenge arises from the noisy gradients prevalent at the onset of training, which are inadequate for guiding the fine-tuning of the encoder effectively. To mitigate this issue, we introduce an auxiliary loss. This loss uses a more coarse-grained objective,
assigning instances to \emph{clusters} of labels, where scores for each cluster are calculated
using a dense classification layer. This auxiliary component stabilizes the gradient flow and enhances the encoder's adaptability during the critical early phases of training and is turned off during later epochs to not interfere with the main task.
Figure \ref{fig:intro-figure} illustrates the architectural changes that ensure
good training performance at different label space sizes sparsity levels.

To materialize actual memory savings, we propose \textsc{Spartex}, which uses semi-structured sparsity~\citep{lasby2024dynamic,schultheis2023towards} with a
fixed fan-in constraint, together
with magnitude-based pruning and random regrowth
(SET~\citep{mocanu2018scalable}), which does not require any additional memory
buffers. 
In our experiments, we show that \textsc{Spartex} achieves a 3.4-fold reduction of GPU memory requirements
from 46.3 to \SI{13.5}{\gibi
\byte} for training on the Amazon-3M~\citep{Bhatia16} dataset, 
with only an approximately 3\% reduction in  predictive performance. In comparison, a na\"ive
parameter reduction using a bottleneck layer at 
the same memory budget decreases precision by about 6\%.

Our primary contributions are as follows:
\begin{itemize}
\item \textbf{Enhancements in training efficiency:} We propose novel modifications to the conventional DST framework that significantly curtail training durations while delivering  competitive performance metrics when benchmarked against dense model baselines and other specialized XMC methodologies. These enhancements are pivotal in demonstrating DST's scalability and efficiency to large label spaces.
\item \textbf{Optimized hardware utilization:} We provide PyTorch bindings for custom CUDA kernels\footnote{Code is available at https://github.com/xmc-aalto/NeurIPS24-dst} which enable a streamlined integration of memory-efficient sparse training into an existing
XMC pipeline. This implementation enables the deployment of our training methodologies on conventional, commercially available hardware, thus democratizing access to state-of-the-art XMC model training.

\item \textbf{Robustness to Label distribution challenges:} Our empirical results demonstrate that the DST framework, as adapted and optimized by our modifications, can effectively manage datasets characterized by label imbalances and the presence of missing labels, with minimal performance degradation for tail labels.
\end{itemize}

\section{Dynamic Sparse Training for Extreme Multi-label Classification}

\subsection{Background} 
\paragraph{Problem setup} Given a multi-label training dataset with \(N\) samples, \(\mathcal{D} = \{(x_i, P_i)_{i=1}^{N}\}\), where \(L\) represents the total number of labels, and \(P_i \subset [L]\) denotes a subset of relevant labels associated with the data point \(x_i\in \chi\).
Typically, the instances are text based, such as the contents of a Wikipedia article~\citep{Bhatia16} or the title of a product
on Amazon~\citep{mcauley2013hidden} with labels corresponding to Wikipedia categories and frequently bought together products, respectively, for example.
Traditional XMC methods used to handle labels the same way as is typically done in other fields, as featureless integers.

However, the labels themselves usually carry some information, e.g., a textual representation, as the  following examples,
taken from (i) LF-AmazonTitles-131K (recommend related products given a product name)
and (ii) LF-WikiTitles-500K (predict relevant categories, given the title of a Wikipedia page) illustrate:

\emph{Example~1:} For \underline{\textit{``Nintendo Land''}} on Amazon, we have available: \textit{Mario Tennis Ultra Smash(Nintendo Wii U)} $\vert$ \textit{Star Fox Zero (Nintendo Wii U)}, as the recommended products. 

\emph{Example~2:} For the \underline{\textit{``2024 United States presidential election''}} Wikipedia page, we have the available categories: \textit{Joe Biden} $\vert$ \textit{Joe Biden 2024 presidential campaign} $\vert$ \textit{Donald Trump} $\vert$ \textit{Donald Trump 2024 presidential campaign} $\vert$ \textit{Kamala Harris} $\vert$ \textit{November 2024 events in the United States}.

Consequently, more recent XMC approaches have started to take these label features into account to alleviate
the data scarcity problems~\citep{dahiya2021siamesexml,mittal2021eclare}.

\paragraph{XMC and DST} XMC models are typically comprised of two main components: (i) an encoder \(\mathcal{E}_\theta : \chi \rightarrow \mathbb{R}^d\), which embeds data points into a \(d\)-dimensional real space, primarily utilizing a transformer architecture~\citep{vaswani2017attention} and (ii) A One-vs-All classifier \(W = \{w_l\}_{l \in [L]}\), where \(w_l\) denotes the classifier for label \(l\), integrated as the last layer of the neural network in end-to-end training settings.
In a typical DST scenario, one would sparsify the language model used as the encoder,
potentially even leaving the classifier fully dense~\citep{kurtic2024prune}. However,
in XMC, most of the networks weights are in the classifier layer, so in order to achieve
a reduction in memory consumption, its weight matrix \(W_s = \{w_{l}^{s}\}_{l \in [L]}\) 
\emph{must} be sparsified.

This sparse layer $W_s$ is then periodically updated in a prune-regrow-train loop,
that is, every $\Delta T$ steps, a fraction of active weights is pruned and the
same number of inactive weights are regrown. The updated sparse topology is then trained with regular
gradient descent for the next $\Delta T$ steps. There are many possible choices for pruning and regrowth
criteria~\citep{hoefler2021sparsity}; to keep memory consumption low, however,
we need to choose a method that does not require auxiliary buffers proportional to
the size of the dense layer. This excludes methods such as requiring second-order information~\citep{hassibi1992second},
or tracking of dense gradients or other per-weight information~\citep{dai2019nest,dettmers2019sparse,Lin2020Dynamic}.
\citet{evci2020rigging} argue that RigL only needs dense gradients in an ephemeral capacity --- they can be discarded
as soon as the regrowth step for the current layer is done, but before the regrow step of the next layer is started ---
but in the XMC setup, the prohibitively large memory consumption arises already from a single layer.
Therefore, we select magnitude-based pruning and random regrowth~\citep{mocanu2018scalable}. Magnitude-based pruning
has been shown to be a remarkably strong baseline~\citep{nowak2024fantastic}.

However, to actually achieve efficient training with these algorithms in the XMC setting,
several challenges need to be overcome as discussed below.



\subsection{Memory-Efficient Training: Fixed Fan-In Sparse Layer}
Unstructured sparsity is notoriously difficult to speed-up on GPUs~\cite{gale2020sparse}, and
consequently most DST studies simulate sparsity by means of a binary mask~\citep{curci2021truly,lee2023jaxpruner}.
On the other hand, highly structured sparsity, such as 2:4 sparsity~\citep{castro2023venom}, enjoys hardware acceleration and memory reduction~\citep{mishra2021accelerating}, but may result in deteriorated model accuracy compared to unstructured sparsity~\citep{liu2023lessonslearnednewsparseland}.
As a compromise, semi-structured sparsity~\citep{muralidharan2023uniform,lasby2024dynamic,schultheis2023towards} imposes a fixed fan-in to each neuron. This eliminates work imbalances between different neurons, leading to an efficient and simple storage format for sparse weights, where each sparse weight needs only a single integer index, resulting in ELLPACK format~\citep{rice1985solving} without any padding.
For 32-bit floating point weights with 16-bit indices (i.e., at most 65k features in the embedding layer), this leads to a 50\% storage overhead for sparse weights;
however, for training, gradient and two momentum terms are needed, which share the same indexing structure, reducing the effective overhead to just 12.5\%.

While fixed fan-in provides substantial speed-ups for the forward pass, due to the transposed matrix multiplication required for gradients, it does not give any direct benefits for the backwards pass.
Fortunately, when used for the classification matrix, the backpropagated gradient is the loss derivative, which will exhibit high \emph{activation sparsity} if the loss function is hinge-like~\citep{schultheis2023towards}.
In the enormous label space of XMC, for each instance only a small subset of labels will be hard negatives. The rest will be easily classified as true negatives, and not contribute to the backward pass.

As additional measures to keep the memory consumption low, we enable \texttt{torch.amp} automatic mixed-precision training~\citep{micikevicius2018mixed}
and activation checkpointing~\citep{chen2016training} for the BERT encoder.


\subsection{Improved Gradient Flow: Auxiliary Objective}
We find that, despite using a fully dense network, training the encoder using gradients
backpropagated from a sparse classification layer requires more optimization steps to converge
compared with to a dense classification layer.
This compounds with
the already-increased number of epochs required for DST~\citep{evci2020rigging,liu2021we}, further increasing
the training duration of end-to-end XMC training, which requires longer
training than comparable modularized or shortlisting-based methods~\citep{jain2023renee}.
Furthermore, the intermediate activations in the transformer-based encoder also
take up a considerable amount of GPU memory, so to meet memory budgets, we may need
to switch to smaller batch sizes or employ activation checkpointing, increasing
the per-step time.

Therefore, we need to improve gradient flow through the sparse
layer. \citet{schultheis2023towards} inserted a large intermediate layer preceding the actual classifier to achieve significant improvements in performance.
While this method is sufficient to achieve good 
results with fixed encodings, we observe that it fails to perform well 
if the encoder is a trainable transformer~\citep{devlin2018bert,sanh2019distilbert}
for label spaces with more than one hundred thousand elements, particularly for 
high sparsity levels.
Therefore, we instead propose to use the label shortlisting task that is typical
for XMC pipelines as an auxiliary objective to generate informative gradients
for the encoder.

\paragraph{Rethinking the role of clusters and meta-classifier in XMC}
Many prevailing XMC methods, apart from learning the per-label classifiers \(W = \{w_l\}_{l \in [L]}\) for the \textit{extreme task}, also employ a meta-classifier.  
The meta-classifier learns to classify over clusters of similar labels that are created by recursively partitioning the label set into equal parts using, for example, balanced $k$-means clustering \cite{jiang2021lightxml,kharbanda2022cascadexml,zhang2021fast,you2019attentionxml, kharbanda2023inceptionxml}. 
These meta-classifiers are primarily used for label shortlisting or retrieval prior to the final classification or re-ranking at the extreme scale. 
We investigated the impact on the final performance of the extreme task when the labels are randomly assigned to the clusters (instead of the following the k-means objective). 
We observed that such reassignments do not negatively affect the extreme task's performance (detailed of this observation are shown in Appendix~\ref{random_cluster}). 
This leads us to hypothesize that beyond merely shortlisting labels, meta-classifier branch of the XMC training pipelines provides useful gradient signals during encoder training, which is particularly crucial for larger datasets with $\mathcal{O}(10^6)$ labels such as Amazon-670K (Figure~\ref{fig:intro-figure}) and Amazon-3M. 

\begin{wrapfigure}[15]{r}{0.5\textwidth}
    \centering
    \vspace{-1ex}
    \includegraphics[width=0.5\textwidth]{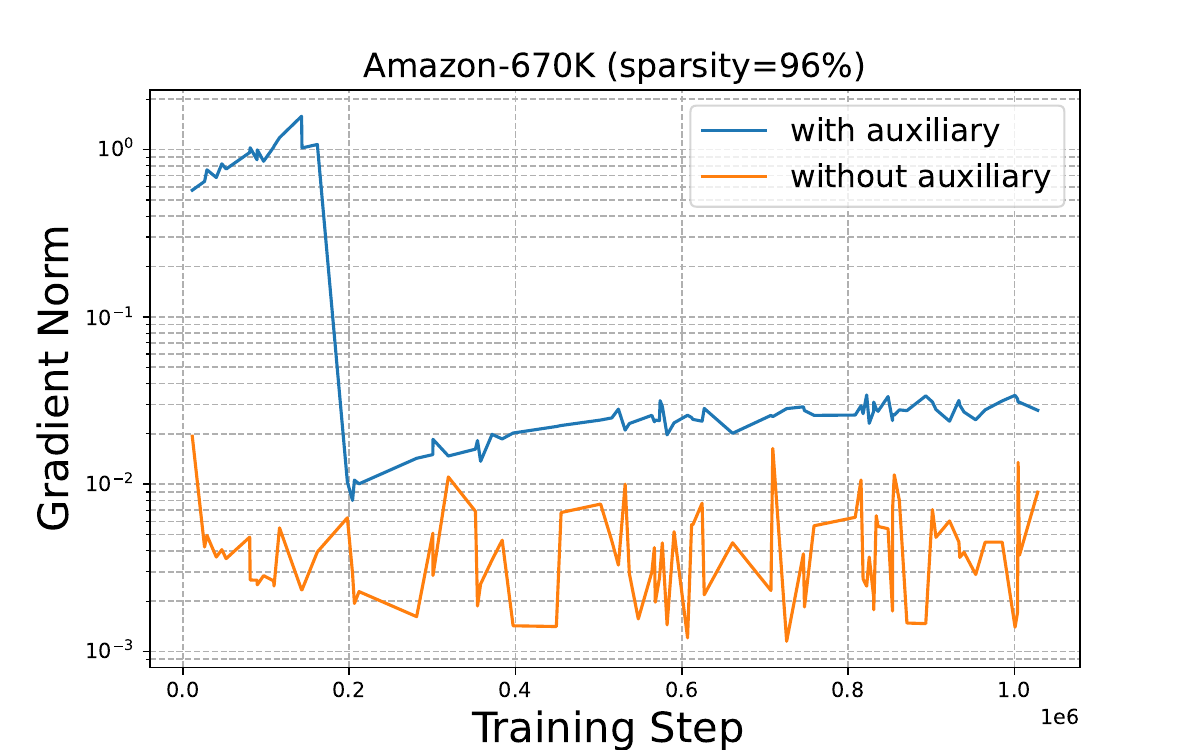}
    \caption{Gradient Flow of the encoder during training with and without Auxiliary Objective.}
    \label{grad_flow}    
\end{wrapfigure}


\paragraph{Auxiliary objective and DST convergence for XMC} Towards addressing the challenge of gradient instability, we augment our training pipeline with an additional meta-classifier branch which aids gradient information during back-propagation. 
This is especially useful during the initial training phase where the fixed-fan-in sparse layer tends to encounter difficulties. 
Importantly, in our model the output layer operates independently of the meta-classifier's outputs, enabling a seamless end-to-end training process.


Although a meta-classifier assists during the initial stages of training, maintaining it throughout the entire training process can deteriorate the encoder's representation quality. This degradation occurs because the task associated with the meta classifiers differs from the final task, yet both share the same encoder. Similar observations have been noted in related studies \cite{kharbanda2022cascadexml}. To address this issue, we implement a decaying scaling factor on the auxiliary loss, gradually reducing its influence as training progresses. 

The impact of the auxiliary branch on the norm of the gradient in the encoder is demonstrated in Figure \ref{grad_flow} for the Amazon-670K dataset.
The larger gradient signal speeds up initial learning, but it is misaligned with the true objective, so is gradually turned off at around 200k steps. Furthermore, the improvement in prediction performance, as reflected in Figure \ref{fig:intro-figure} (right panel), reinforces the quality of gradient as compared to the training pipeline without the auxiliary objective.

\section{Experiments and discussion}
\label{experiment_and_discussion}

\subsection{Datasets}
In this study, we evaluate our proposed modifications of DST under the extreme classification setting across a diverse set of large-scale datasets, including Wiki10-31K~\citep{zubiaga2012enhancing}, Wiki-500K~\citep{Bhatia16}, Amazon-670K~\citep{mcauley2013hidden}, and Amazon-3M~\citep{mcauley2015inferring}. 
The datasets are publicly available at the Extreme Classification Repository\footnote{\url{http://manikvarma.org/downloads/XC/XMLRepository.html}}.  
These were selected due to their inherent complexity and the challenges posed by their long-tailed label distributions, which are emblematic of real-world data scenarios and test the robustness of DST methodologies. Further validation of our approach is conducted using the datasets LF-AmazonTitles-131K and LF-WikiSeeAlso-320K. These datasets are particularly relevant due to their augmentation with rich label metadata and their concise text formats, traits that have gained considerable traction in the XMC research community of late. The datasets' detailed statistical profiles are delineated in Table \ref{tab:dataset_stats}.\looseness=-1

\begin{table}[tbp]
  \caption{Statistics of XMC Datasets with and without Label Features. This table presents a comparison across various datasets, detailing the total number of training instances ($N$), unique labels ($L$), number of test instances ($N'$), average label count per instance ($\overline{L}$), and average data points per label ($\hat{L}$).}
  \label{tab:dataset_stats}
  \centering
  \begin{tabular}{
    @{}l
    S[table-format=7.0,group-separator={,}]
    S[table-format=6.0,group-separator={,}]
    S[table-format=7.0,group-separator={,}]
    S[table-format=2.2]
    S[table-format=3.2]@{}
  }
    \toprule
    \textbf{Dataset} & {$N$} & {$L$} & {\textbf{$N'$}} & {\textbf{$\overline{L}$}} & {\textbf{$\hat{L}$}} \\
    \midrule
    \multicolumn{6}{c}{\textbf{Datasets without Label Features}} \\
    \midrule
    Wiki10-31K          & 14146   & 30938  & 6616    & 18.64 & 48.52 \\
    Wiki-500K           & 1779881 & 501070 & 769421  & 4.75  & 16.86 \\
    Amazon-670K         & 490449  & 670091 & 153025  & 5.45  & 3.99 \\
    Amazon-3M           & 1717899 & 2812281& 742507  & 36.17 & 31.64 \\
    \midrule
    \multicolumn{6}{c}{\textbf{Datasets with Label Features}} \\
    \midrule
    LF-AmazonTitles-131K & 294805  & 131073 & 134835  & 5.15  & 2.29 \\
    LF-WikiSeeAlso-320K & 693082 & 312330 & 177515 & 4.67  & 2.11 \\

    \bottomrule
  \end{tabular}
\end{table}
\subsection{Baselines and evaluation metrics}
To ensure a comprehensive and fair evaluation of our proposed DST methodologies applied to XMC problems, we compare our proposed framework, \textsc{Spartex}, across three principal categories of baseline methods:
\begin{enumerate}
    \item \textbf{Dense Models:} Consistent with traditional DST evaluations, we compare the performance of our sparse models against their dense counterparts. 
    \item \textbf{Dense Models with bottleneck layer:} This category (referred to as Dense BN in Table \ref{main-table}) includes dense models with the same number of parameters as our proposed DST method by having a bottleneck layer with the same dimensionality as the FFI size. This ensures that comparisons focus on the impact of sparsity rather than differences in model size or capacity.
    \item \textbf{XMC Methods:} For datasets devoid of label features, we benchmark against the latest transformer-based models such as \textsc{CascadeXML} \cite{kharbanda2022cascadexml}, \textsc{LightXML} \cite{jiang2021lightxml}, and \textsc{XR-Transformer}. For datasets that incorporate label features, our comparison includes leading Siamese methods like \textsc{SiameseXML} \cite{dahiya2021siamesexml} and \textsc{NGAME}\cite{dahiya2023ngame}, as well as other relevant transformer-based approaches.
    
\end{enumerate}
    Notably, \textsc{Renee} \cite{jain2023renee} qualifies as both a dense model and a state-of-the-art XMC method. However, in some instances, \textsc{Renee} employs larger encoders (e.g., Roberta-Large \cite{liu2019roberta}). To maintain consistency and fairness in our evaluations, we exclude configurations employing larger encoders from this analysis. For conceptual validation, we used \textsc{RigL}\cite{evci2020rigging} on datasets with label spaces up to 670K.

As is standard in XMC literature, we compare the methods on metrics which only consider prediction at top-k slots. This includes : Precision@k and its propensity-scored variant (which is more sensitive to performance on tail-labels). The details of these metrics are given in Appendix \ref{metrics}.

\subsection{Empirical performance}
Table \ref{main-table} presents our primary results on the datasets, compared with the aforementioned baselines. The performance metrics for XMC baselines are reported from their original papers. However, for peak memory consumption, we re-ran these baselines in half precision with the same batch size, as all baselines are also evaluated in half precision.  Following DST protocols, we extended the training duration for \textsc{RigL}, Dense Bottleneck, and our method to twice the number of steps used for dense models. Certain baselines (referred to as OOM) could not scale to the Amazon-3M dataset. Our results demonstrate that our method significantly reduces memory usage while maintaining competitive performance. On the Amazon-3M dataset, our approach delivers comparable performance to dense models while achieving a 3.4-fold reduction in memory usage and a 5-fold reduction compared to the XMC baseline. Furthermore, within the memory-efficient model regime, our method consistently outperforms the Dense Bottleneck model. To further validate the robustness of our approach, we evaluated it on the label features datasets, as shown in Table \ref{lf-table}. Notably, as the label space size increases, we need to adjust to a comparatively lower sparsity to maintain performance, discussed in detail in subsequent sections.

\begin{table}[t]
\caption{Comparison with different methods. Comparing our sparse model against its dense counterpart and with state-of-the-art XMC methods on Wiki10-31K, Wiki-500K, Amazon-670K and Amazon-3M datasets. \(M_\mathrm{tr}(\mathrm{GiB})\) indicates peak GPU memory consumption during training.}
\label{main-table}
\begin{center}
\begin{small}
\begin{tabular}{@{}l|c|cccc|c|cccc@{}}
\toprule
Method & \makecell{Sparsity \\ (\%)} & P@1 & P@3 & P@5 & \makecell{\(M_\mathrm{tr}\)\\\((\mathrm{GiB})\)} & \makecell{Sparsity \\ (\%)} & P@1 & P@3 & P@5 & \makecell{\(M_\mathrm{tr}\)\\\((GiB)\)} \\
\midrule \midrule
& & \multicolumn{4}{c|}{Wiki10-31K} & & \multicolumn{4}{c}{Wiki-500K} \\
\midrule
\textsc{AttentionXML} & - & 87.1 & 77.8 & 68.8 & 8.2 & - & 75.1 & 56.5 & 44.4 & 13.1 \\
\textsc{LightXML} & - & 87.8 & 77.3 & 68.0 & 16.5 & - & 76.2 & 57.2 & 44.1 & 14.6 \\
\textsc{CascadeXML} & - & 88.4 & 78.3 & 68.9 & 8.2 & - & 77.0 & 58.3 & 45.1 & 18.8 \\
\midrule
\textsc{Dense} & - & 87.8 & 77.2 & 68.1 & 2.5 & - & 78.5 & 59.2 & 45.6 & 9 \\
\midrule
\textsc{Dense BN} & - & 86.7 & 76.3 & 66.0 & 2.1 & - & 73.8 & 55.1 & 42.0 & 4.3 \\
\textsc{RigL}  & 92 & 87.7 & 77.3 & 67.7 & 2.6 & 83 & 74.5 & 54.7 & 41.8 & 9.7  \\
\textsc{Spartex} & 92 & 88.6 & 77.7 & 67.4 & \textbf{2.1} & 83 & 76.7 & 57.8 & 44.5 & \textbf{4.1} \\
\midrule \midrule
& & \multicolumn{4}{c|}{Amazon-670K} & & \multicolumn{4}{c}{Amazon-3M} \\
\midrule
\textsc{AttentionXML} & - & 45.7 & 40.7 & 36.9 & 10.7 & - & 49.1 & 46.0 & 43.9 & 71.2 \\
\textsc{LightXML} & - & 47.1 & 42.0 & 38.2 & 11.2 & - & - & - &  & OOM \\
\textsc{CascadeXML} & - & 48.5 & 43.7 & 40.0 & 18.3 & - & 51.3 & 49.0 & 46.9 & 87.0 \\
\midrule
\textsc{Dense} & - & 49.8 & 44.2 & 40.1 & 11.5 & - & 53.4 & 50.6 & 48.5 & 46.3 \\
\midrule
\textsc{Dense BN} & - & 44.5 & 39.7 & 36.1 & 4.0 & - & 47.0 & 44.6 & 42.7 & 13.1  \\
\textsc{RigL} & 83 & 45.2 & 38.7 & 36.0 & 12.4 & 83 & - & - & - & OOM \\
\textsc{Spartex} & 83 & 47.1 & 41.8 & 38.0 & \textbf{3.7} & 83 & 50.2 & 47.1 & 44.8 & 13.5 \\
\bottomrule
\end{tabular}
\end{small}
\end{center}
\end{table}

\begin{table}[b]
\caption{Comparison on label feature datasets LF-AmazonTitles-131K and LF-WikiSeeAlso-320K. \(M_\mathrm{tr}(\mathrm{GiB})\) indicates peak GPU memory usage during training.}
\label{lf-table}
\begin{center}
\begin{small}
\begin{tabular}{@{}l|c|cccc|c|cccc@{}}
\toprule
Method & \makecell{Sparsity \\ (\%)} & P@1 & P@3 & P@5 & \makecell{\(M_\mathrm{tr}\)\\\((\mathrm{GiB})\)} & \makecell{Sparsity \\ (\%)} & P@1 & P@3 & P@5 & \makecell{\(M_\mathrm{tr}\)\\\((GiB)\)} \\
\midrule \midrule
& & \multicolumn{4}{c|}{LF-AmazonTitles-131K} & & \multicolumn{4}{c}{LF-WikiSeeAlso-320K} \\
\midrule
\textsc{LightXML} & - &  35.6 & 24.2 & 17.5 & 11.6 & - & 34.5 & 22.3 & 16.8 & 13.5 \\
\textsc{ECLARE} & - & 40.7 & 27.5 & 19.9 & 8.8 & - &  40.6 & 26.9 & 20.1 & 10.2  \\
\textsc{SiameseXML} & - &  41.4 & 27.9 & 21.2 & 7.1 & - & 42.2 & 28.1 & 21.4 & 8.9 \\
\textsc{NGAME} & - & 46 & 30.3 & 21.5 & 9.0 & - & 47.7 & 31.6 & 23.6 & 19.3 \\
\textsc{DEXML} & - & 42.5 & - & 20.6 & 30.2 & - & 46.1 & 29.9 & 22.3 & 56.1 \\
\midrule
\textsc{Renee} (Dense) & - &  46.1 & 30.8 & 22 & 3.0  & - & 47.9 & 31.9 & 24.1 & 9.1 \\
\midrule
\textsc{Dense BN} & - & 39.2 & 25.7 & 18.2 & 2.2 & - & 44.5 & 28.4 & 21.5 & 6.0  \\
\textsc{RigL} & 83 & 43.0 & 28.6 & 20.4 & 3.2 & 67 & 44.9 & 29.0 & 21.7 & 9.2  \\
\textsc{Spartex} &  83 &  44.5 & 29.8 & 21.3 & \textbf{2.2} & 67 & 46.0 & 29.9 & 22.1 & 6.1  \\
\bottomrule
\end{tabular}
\end{small}
\end{center}
\end{table}

\begin{figure}
    \centering
    \includegraphics[width = \textwidth]{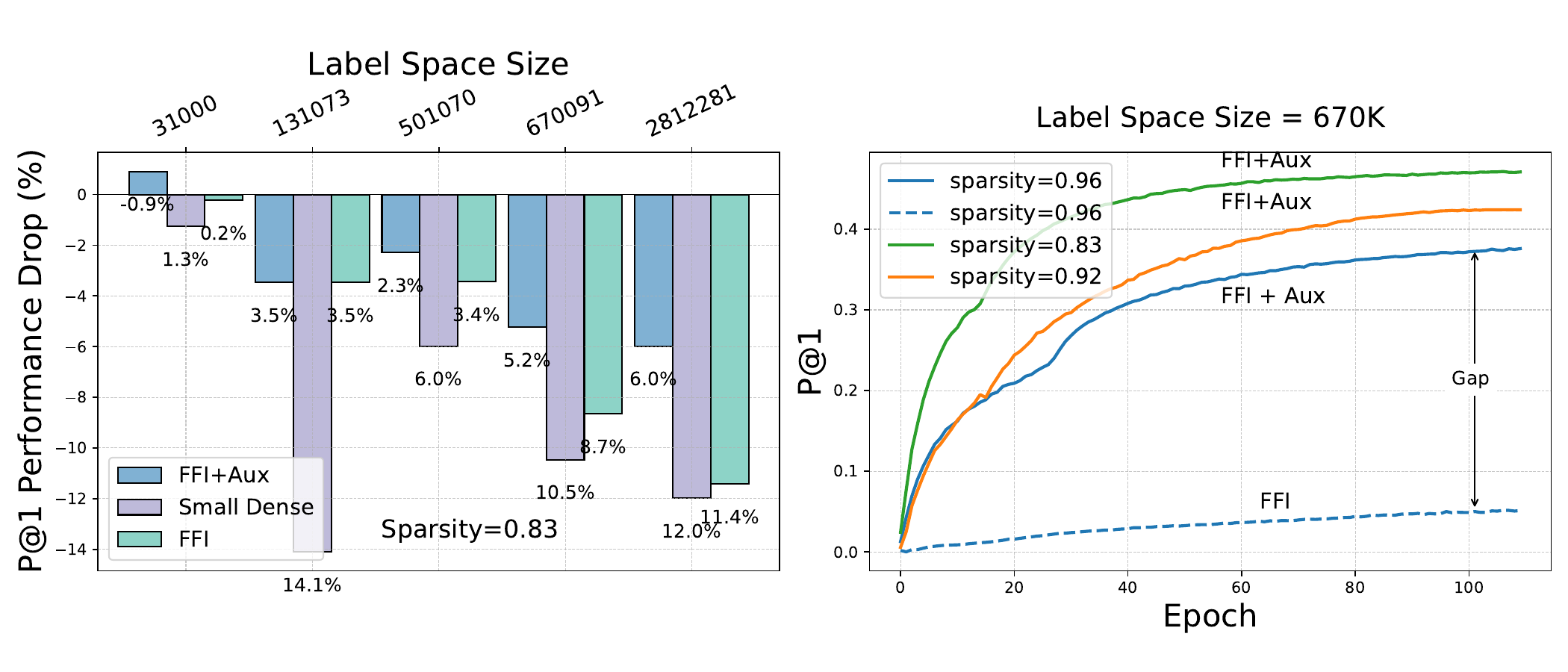}
    \caption{\textbf{left:} Comparison of performance declines as the size of the label space increases, given a fixed sparsity. \textbf{right:} Performance of our model at different epochs, across various sparsity ratios. }
    \label{labelsize_sparsity}    
\end{figure}

\subsection{Adapting to increased sparsity and label size: the role of auxiliary objective} The DST approach is widely recognized to be problematic when dealing with high sparsity levels ($\geq90\%$). This is also apparent in our experiment and can be observed in Figure \ref{labelsize_sparsity} (right) when the label space size is constant. Our findings indicate that incorporating an auxiliary objective significantly aids in maintaining performance, particularly in the high sparsity regime. Conversely, at lower sparsity levels ($\leq 67\%$), the benefit of the auxiliary objective diminishes.
In the context of XMC problems, the performance of DST degrades as the label space size increases. Figure \ref{labelsize_sparsity} (left) depicts the performance degradation of our approach relative to a dense baseline across datasets with increasing label space sizes: 31K, 131K, 500K, 670K, and 3M (detailed in the Table \ref{tab:dataset_stats}), all evaluated at 83\% sparsity. Interestingly, for the wiki31K dataset, we observe a performance improvement, potentially due to the lower number of training samples relative to the label space size. Compared to other methods with equivalent memory requirements, our approach demonstrates superior performance retention at larger label space sizes.

%


%
\subsection{Effect of Rewiring Interval } 
\begin{figure}[b]
  \centering
  \includegraphics[width=\textwidth]{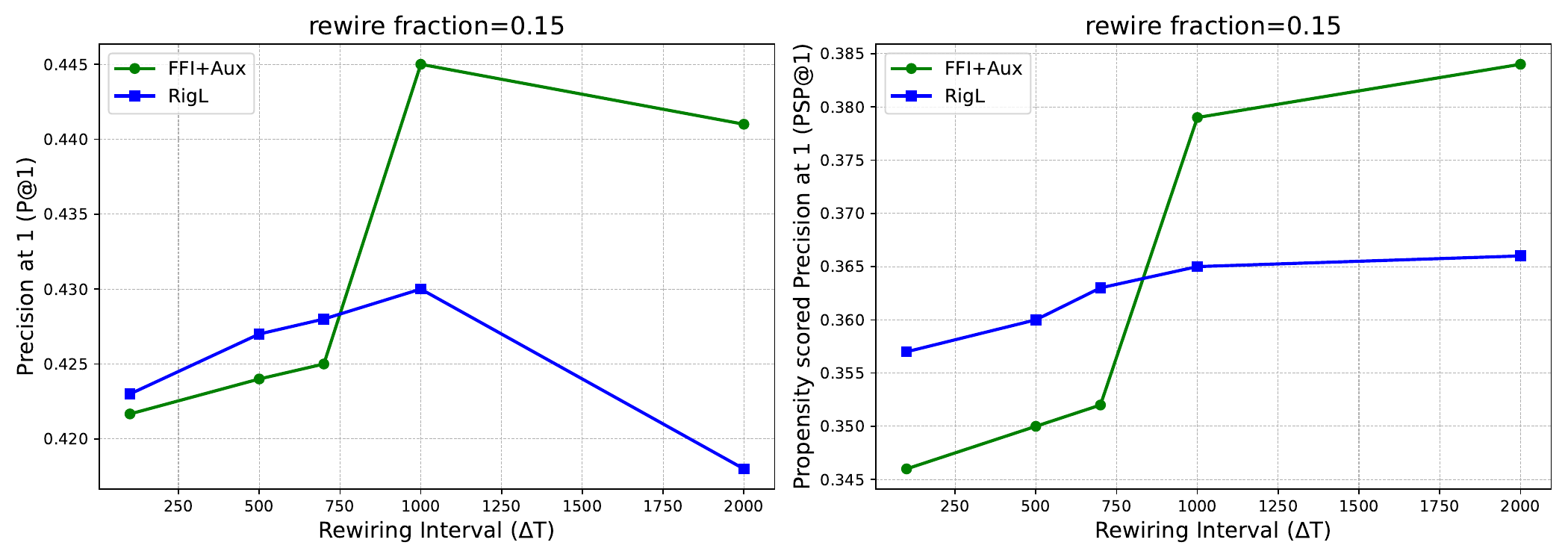}
  \caption{Effect of rewiring interval on final performance for Precision@1 \textbf{(left)} and propensity-scored Precision@1 \textbf{(right)} in the LF-AmazonTitles-131K dataset.}
  \label{rewire_interval}
\end{figure}

The rewiring interval is crucial for balancing the trade-off between under-exploration and unreliable exploration. In XMC problems, tail label performance is particularly significant due to its application domain. The rewiring interval directly influences how frequently each parameter topology encounters tail label examples before updates. In this section, we focus on assessing the performance impact of various rewiring intervals, including their effect on tail labels. We conducted experiments on the LF-AmazonTitles-131K dataset using rewiring intervals $ \Delta T \in [100,500,700,1000,2000] $. The corresponding results for P@1 and PSP@1 metrics are illustrated in Figures \ref{rewire_interval} left and right, respectively, with a fixed rewiring fraction of $0.15$. Our findings reveal that both P@1 and PSP@1 improve as the interval increases up to a certain point. Interestingly, while P@1 shows a decline beyond this threshold, PSP@1 continues to rise. This divergence suggests that larger rewiring intervals, despite potentially limiting the diversity of topology exploration, provide each topology sufficient exposure to more tail labels, thereby improving model performance in handling rare categories.

\begin{table}[tb]
\caption{Propensity-Scored Precision (PSP) comparison of our sparse model with its dense counterpart and state-of-the-art XMC methods on the Wiki10-31K, Wiki-500K, Amazon-670K, and Amazon-3M datasets. The same sparsity levels as mentioned in previous tables are used. }
\label{psp-table}
\centering
\begin{small}
\begin{tabular}{@{}l|ccc|ccc@{}}
\toprule
Method &  PSP@1 & PSP@3 & PSP@5 &  PSP@1 & PSP@3 & PSP@5  \\
\midrule \midrule
&  \multicolumn{3}{c|}{Wiki10-31K}  & \multicolumn{3}{c}{Wiki-500K} \\
\midrule
\textsc{AttentionXML} &  16.2 & 17.1 & 17.9 &   30.1 & 37.3 & 41.7  \\
\textsc{CascadeXML} &  13.2 & 14.7 & 16.1 &  31.3 & 39.4 & 43.3  \\
\textsc{Dense} &  10.6 & 12.6 & 13.9 &  32.5 & 41.0 & 44.9  \\
\textsc{Dense BN}  & 12.4 & 14.5 & 16.0   & 28.5 & 36.5 & 40.1  \\
\textsc{Spartex} &  13.2 & 15.1 & 16.4 &  31.9 & 39.8 & 43.4  \\
\midrule
&  \multicolumn{3}{c|}{Amazon-670K}  & \multicolumn{3}{c}{Amazon-3M} \\
\midrule
\textsc{AttentionXML} &  29.3 & 32.4 & 35.1 &   15.5 & 18.5 & 20.6  \\
\textsc{CascadeXML} &  30.2 & 34.9 & 38.8 &  - & - & -  \\
\textsc{Dense} &  33.3 & 36.9 & 39.9 &  15.6 & 19.0 & 21.5  \\
\textsc{Dense BN}  & 26.9 & 30.9 & 34.5   & 13.5 & 16.4 & 18.6  \\
\textsc{Spartex} &  29.9 & 33.3 & 36.4 &  14.3 & 17.2 & 19.4  \\
\bottomrule
\end{tabular}
\end{small}
\end{table}

\subsection{Performance on Tail Labels } Table \ref{psp-table} presents a comparison of Propensity-Scored Precision (PSP) for various Extreme Multi-label Classification (XMC) models, including \textsc{AttentionXML} \cite{you2019attentionxml}, \textsc{CascadeXML} \cite{kharbanda2022cascadexml}, Dense, Dense Bottleneck, and our proposed method, across four benchmark datasets: Wiki10-31K, Wiki-500K, Amazon-670K, and Amazon-3M. For the Wiki10-31K dataset, our model achieves PSP@1 of 13.2, PSP@3 of 15.1, and PSP@5 of 16.4, surpassing the Dense model. On the Wiki-500K dataset, our method records a PSP@1 of 31.9, outperforming other XMC models and closely trailing the top-performing approaches. These findings underscore our model's consistent performance across varied datasets, frequently exceeding or closely competing with XMC and Dense benchmarks. It's noteworthy that the performance of \textsc{AttentionXML} on Wiki10-31K is attributed to its utilization of an LSTM encoder, which is particularly advantageous given the dataset's smaller number of training samples relative to its label space. This configuration also explains our model's superior performance compared to the Dense model, which incorporates a form of regularization. In comparisons involving memory efficiency, our approach significantly surpasses the same-capacity Dense Bottleneck model, demonstrating its suitability in resource-constrained settings where tail-label performance is critical.

\begin{table}[b]
\caption{Comparison of Fan-in (sparsity) effects on model performance and memory usage for Amazon-670K dataset.}
\label{fan-in-table}
\begin{center}
\begin{small}
\begin{tabular}{@{}c|c|cccc|cc@{}}
\toprule
\makecell{Fan in  (sparsity)} & Auxiliary Loss & P@1 & P@3 & P@5 & \makecell{\(M_\mathrm{tr}\) \((\mathrm{GiB})\)} & \makecell{Epoch Time \\ (mins)} & \makecell{Inference Time \\ (ms)} \\
\midrule \midrule
384 (50\%) & No & 49.0 & 43.5 & 39.5 & 7.03 & 18:13 & 10.2 \\
384 (50\%) & Yes & \textbf{49.2} & \textbf{43.7} & \textbf{39.6} & 7.13 & 18:19 & 10.2 \\
256 (67\%) & No & 47.0 & 41.6 & 37.7 & 5.27 & 15:45 & 9.18 \\
256 (67\%) & Yes & 47.6 & 42.3 & 38.4 & 5.36 & 15:50 & 9.18 \\
128 (83\%) & No & 45.6 & 40.2 & 36.3 & 3.60 & 13:20 & 8.54 \\
128 (83\%) & Yes & 47.1 & 41.8 & 38.0 & 3.70 & 13:23 & 8.54 \\
64 (92\%) & No & 30.7 & 26.5 & 23.6 & 2.88 & 12:12 & 8.14 \\
64 (92\%) & Yes & 42.3 & 37.2 & 33.3 & 2.97 & 12:13 & 8.14 \\
32 (96\%) & No & 5.5 & 5.0 & 4.6 & \textbf{2.52} & \textbf{11:36} & \textbf{7.94} \\
32 (96\%) & Yes & 38.4 & 33.8 & 30.4 & 2.61 & 11:38 & \textbf{7.94} \\
\bottomrule
\end{tabular}
\end{small}
\end{center}
\end{table}

\subsection{Impact of Varying Sparsity Levels } Table \ref{fan-in-table} illustrates the impact of varying sparsity levels (ranging from 50\% to 96\%) in conjunction with the use of auxiliary loss for Amazon-670K dataset. As sparsity levels increase, there are benefits in memory usage, training time, and inference time; however, performance metrics simultaneously decline. Additionally, the importance of auxiliary loss becomes particularly significant at higher sparsity levels.

\subsection{Sensitivity to Auxiliary Loss cut-off epochs} We employ auxiliary loss with an initial scalar weight that decays until a specified cut-off epoch. Table \ref{auxiliary-cutoff-table} illustrates the model's final performance at various cut-off epochs for two sparsity levels. A value of 0 (No aux) indicates the absence of auxiliary loss, while 'No cut-off' signifies its application throughout training. Our analysis reveals that prolonging the auxiliary loss beyond optimal cut-off epochs adversely affects performance for both sparsity levels. Notably, maintaining the auxiliary loss throughout training leads to performance deterioration, resulting in scores lower than those achieved without its use.

\begin{table}[]
\caption{Comparison of Auxiliary loss cut-off epoch effects on model performance for different Fan-in (sparsity) levels.}
\label{auxiliary-cutoff-table}
\begin{center}
\begin{small}
\begin{tabular}{@{}c|ccc|c|ccc@{}}
\toprule
\multicolumn{4}{c|}{Fan in: 128 (83\% Sparsity)} & \multicolumn{4}{c}{Fan in: 64 (92\% Sparsity)} \\
\midrule \midrule
\makecell[l]{Auxiliary cut-off epoch} & P@1 & P@3 & P@5 & \makecell[l]{Auxiliary cut-off epoch} & P@1 & P@3 & P@5 \\
\midrule 
    0 (No aux) & 45.6 & 40.2 & 36.3 & 0 (No aux) & 30.7 & 26.5 & 23.6 \\
15 & \textbf{47.1} & \textbf{41.8} & \textbf{38.0} & 15 & \textbf{42.3} & \textbf{37.2} & \textbf{33.3} \\
30 & 46.6 & 41.1 & 37.1 & 30 & 32.8 & 28.3 & 25.2 \\
60 & 45.9 & 40.6 & 36.6 & 60 & 32.5 & 28.0 & 24.9 \\
90 & 44.6 & 39.7 & 35.9 & 90 & 31.3 & 27.0 & 23.9 \\
\makecell[l]{No cut-off  (full training)} & 42.1 & 37.4 & 33.7 & \makecell[l]{No cut-off  (full training)} & 22.7 & 17.7 & 14.4 \\
\bottomrule
\end{tabular}
\end{small}
\end{center}
\end{table}

\subsection{DST with Fixed Embedding vs End-to-End Training}

\begin{table}
\caption{Performance comparison between fixed CascadeXML \cite{kharbanda2022cascadexml} embeddings and end-to-end training with DST on Wiki-500K and Amazon-670K datasets.}
\centering
\small
\begin{tabular}{l@{\hspace{8pt}}ccc@{\hspace{16pt}}ccc}
\toprule
& \multicolumn{3}{c}{Wiki-500K} & \multicolumn{3}{c}{Amazon-670K} \\
\cmidrule(r){2-4} \cmidrule(l){5-7}
Method & P@1 & P@3 & P@5 & P@1 & P@3 & P@5 \\
\midrule
Fixed & 73.6 & 54.8 & 42.1 & 42.6 & 37.1 & 33.1 \\
End-to-End & \textbf{76.7} & \textbf{57.8} & \textbf{44.5} & \textbf{47.1} & \textbf{41.8} & \textbf{38.0} \\
\bottomrule
\end{tabular}

\label{tab:fixedvsete_results}
\end{table}
In Table \ref{tab:fixedvsete_results}, we compare the performance of models using fixed  embeddings \cite{schultheis2023towards} with trained end-to-end using DST on the Wiki-500K and Amazon-670K datasets. End-to-end training yields consistent improvements over fixed embeddings across all metrics, with significant gains in P@1 (an increase of 3.1\% on Wiki-500K and 4.5\% on Amazon-670K). These highlight the need of enabling the model to adapt its representations while training for the best possible performance.

\section{Conclusion and future work}

In this paper, we demonstrated the feasibility of DST
for end-to-end training of classifiers with hundreds of thousands of labels.
When appropriately adapted for XMC problems with fixed fan-in sparsity and an auxiliary objective, DST offers a significant reduction in peak memory usage while delivering superior performance compared to bottleneck-based weight reduction. 
It is anticipated that the Python bindings of the CUDA kernels will be useful for the research community in making their existing and forthcoming deep XMC pipelines more memory efficient.
We hope that our work will enable further research towards developing techniques which could be (i) combined with explicit negative mining strategies, and (ii) used in conjunction with larger transformer encoders such as Roberta-large.


\section{Limitations and societal impact}
For XMC tasks, this paper attempts to explore the landscape of sparse neural networks, which can be trained on affordable and easily accessible commodity GPUs. 
While the proposed scheme is able to achieve comparable results to state-of-the-art methods at a fraction of GPU memory consumption, it is unable to surpass the baseline with dense last layer on all occasions. The exact relative decline in prediction performance,  when our proposed adaptations of Fixed Fan-In and auxiliary objective are employed, is shown in Figure \ref{labelsize_sparsity}. 

While we do not anticipate any negative societal impact of our work, it is expected that it will further enable the exploration of novel training methodologies for deep networks which are more affordable and easily accessible to a broader research community outside the big technology companies.

\section{Acknowledgements}
We thank Niki Loppi of NVIDIA AI Technology Center Finland for useful discussions on the sparse CUDA kernel implementations. 
YI acknowledges the support of  Alberta Innovates (ALLRP-577350-22, ALLRP-222301502), the Natural Sciences and Engineering Research Council of Canada (RGPIN-2022-03120, DGECR-2022-00358), and Defence Research and Development Canada (DGDND-2022-03120). 
This research was enabled in part by support provided by the Digital Research Alliance of Canada (alliancecan.ca).
RB acknowledges the support of Academy of Finland (Research Council of Finland) via grants 347707 and 348215.
NU acknowledges the support of computational resources provided by the Aalto Science-IT project, and CSC IT Center for Science, Finland.

{
\small

\bibliographystyle{unsrtnat} 
\bibliography{ref} 

}

\newpage

\appendix

\section{Evaluation Metrics} \label{metrics}
To evaluate the performance of our Extreme Multi-label Text Classification (XMC) model, which incorporates Dynamic Sparse Training, we use a set of metrics designed to provide a comprehensive analysis of both overall and label-specific model performance. The primary metrics we employ is Precision at $k$ (P@$k$), which assess the accuracy of the top-$k$ predictions. Additionally, we incorporate Propensity-Scored Precision at $k$ (PSP@$k$), Macro Precision at $k$ (Macro P@$k$) and Macro Recall at $k$ (Macro R@$k$) to gauge the uniformity of the model's effectiveness across the diverse range of labels typical in XMC problems.
\paragraph{Precision at $k$ (P@$k$):} Precision at k is the fundamental metric for evaluating the top-$k$ predictions in XMC applications such as e-commerce product recommendation and document tagging:
\begin{equation}
    P@k(y, \hat{y}) = \frac{1}{k} \sum_{\ell \in \text{top}_k(\hat{y})} y_\ell
\end{equation}
where $y$ is the true label vector, $\hat{y}$ is the predicted score vector, and $ \text{top}_k(\hat{y})$ identifies the indices with the top-$k$ highest predicted scores.

\paragraph{Propensity-Scored Precision at $k$ (PSP@$k$):} Given the long-tailed label distribution in many XMC datasets, PSP@k incorporates a propensity score 
$y_l$ to weight the precision contribution of each label, thereby emphasizing the tail labels' performance:
\begin{equation}
    PSP@k(y, \hat{y}) = \frac{1}{k} \sum_{\ell \in \text{top}_k(\hat{y})} \frac{y_\ell}{p_\ell}
\end{equation}
where $p_l$ corresponds to the propensity score for the label $y_l$ \cite{jain2016extreme}.

\paragraph{Macro Precision at $k$ (Macro P@$k$):} To capture the average precision across all labels and mitigate any label imbalance, Macro Precision at $k$ is used:
\begin{equation}
    \text{Macro } P@k = \frac{1}{L} \sum_{i=1}^L \left(\frac{\sum_{\ell \in \text{top}_k(\hat{y}_i)} y_{i\ell}}{\min(k, \ |\text{top}_k(\hat{y}_i)|)}\right)
\end{equation}


\section{Baselines and Related Works} Despite the Dense and Same Capacity Dense baseline we also compare our approach with different State of the Art XMC methods and DST methods. 

\paragraph{XMC Methods} We compare our method with deep XMC methods with mainly transformer encoder.
\begin{itemize}
    \item \textbf{AttentionXML}\cite{you2019attentionxml}: The model segments labels using a shallow and wide PLT with a depth between 2 and 3, learning a specific context vector for each label to create label-adapted datapoint representations.
    \item \textbf{LightXML} \cite{jiang2021lightxml}: The method employs a transformer encoder to concurrently train both the retriever and ranker, which incorporates dynamic negative sampling to enhance the model's efficacy.
    \item \textbf{XR-Transformer} \cite{zhang2021fast}: XR-Transformer employs a multi-resolution training approach, iteratively training and freezing the transformer before re-clustering and re-training classifiers at various resolutions using fixed features.
    \item \textbf{CascadeXML} \cite{kharbanda2022cascadexml}:  This method separates the feature learning of distinct tasks across various layers of the Probabilistic Label Tree (PLT) and aligns them with corresponding layers of the transformer encoder. 
    \item \textbf{ECLARE} \cite{mittal2021eclare}: This model utilizes label graphs to improve label representations, focusing specifically on enhancing performance for rare labels. The label graph is generated through random walks using the label vectors.
    \item \textbf{SiameseXML} \cite{dahiya2021siamesexml}: This approach combines Siamese networks with one-vs-all classifiers. SiameseXML utilizes multiple ANNS structures to retrieve label shortlists. These shortlisted labels are subsequently ranked based on scores from label-wise one-vs-all classifiers.
    \item \textbf{NGAME} \cite{dahiya2023ngame}: NGAME enhances transformer-based training for extreme classification by introducing a negative mining-aware mini-batching technique, which supports larger batch sizes and accelerates convergence by optimizing the handling of negative samples.
    \item \textbf{Renee} \cite{jain2023renee}: The Renee model employs an integrated end-to-end training approach for extreme classification, using a novel loss shortcut for memory optimization and a hybrid data-model parallel architecture to enhance training efficiency and scalability.
\end{itemize}

\paragraph{DST Methods} Existing DST methods vary in their pruning and growing criteria. Recent studies \cite{nowak2024fantastic} indicate that magnitude-based pruning is effective in DST, while dense weight information is impractical for Extreme Multi-label Classification (XMC). We evaluate key methods on select datasets for conceptual validation.

\begin{itemize}
    \item \textbf{RigL} \cite{evci2020rigging}: RigL uses weight magnitude and dense gradient magnitudes for the pruning and regrowth saliency criteria, respectively. While RigL only needs the dense gradient information during network topology updates, this is a prohibitive requirement in the XMC setting due to the large memory consumption of the final classification layer. RigL learns an unstructured sparse network toplogy, which is challenging to accelerate on GPUs. 
    \item \textbf{Structured DST} \cite{muralidharan2023uniform,lasby2024dynamic,schultheis2023towards}: In contrast to RigL, structured DST methods adds constraints to the learned network topology such that the network is more amenable to acceleration on commodity GPUs. In our case, we employ the fixed fan-in constraint which reduces both the latency and memory consumption of the final classification layer in the XMC setting. Further, since the dense gradient information is not available in the XMC task, we simply randomly regrow weights as per to SET~\citep{mocanu2018scalable} which has proven to be a robust baseline in the DST literature. 
\end{itemize}

\section{Hyperparameter Settings} We present the hyperparameter settings used during training in Table \ref{hyperparameter_training}. For the encoder and classifier, we employ two separate optimizers: AdamW for both components, except in the case of LF-AmazonTitles-131K where Adam and SGD are utilized. All experiments are conducted using half-precision \textit{float16} types, except for Amazon-3M and LF-AmazonTitles-131K, which use the \textit{bfloat16} type. We apply a cosine scheduler with warmup, as specified in the table. The weight decay values are set separately: 0.01 for the encoder and 1.0e-4 for the final classification layer. We use the squared hinge loss function for all datasets except for LF-AmazonTitles-131K, where we use binary cross-entropy (BCE) loss with positive labels.

\begin{table}[ht]
    \centering
    \caption{Hyperparameters of our approach to facilitate reproducibility. "LR" stands for learning rate.}
    \begin{tabular}{c|cccccccc}
    \midrule
        Dataset & Encoder & \makecell{Batch \\Size} & Dropout & Epochs & \makecell{LR \\Encoder}  & \makecell{LR \\Classifier} & Warmup & \makecell{Sequence\\Length}\\
        \midrule
        \makecell{Wiki10\\31K} & \makecell{BERT\\Base} & 32 &  0.5 & 100 & 1.0e-5 & 0.01 & 1000 & 128 \\
        \midrule
        \makecell{LF\\AmazonTtles\\131K} & \makecell{Distil\\BERT} & 512 & 0.7 & 200 & 1.0e-5 & 0.08 & 5000  & 32 \\
        \midrule
        \makecell{Wiki\\500K} & \makecell{BERT\\Base} & 128 & 0.5 & 50 &  5.0e-5 & 0.05 & 1000  & 128 \\
        \midrule
        \makecell{Amazon\\670K} & \makecell{BERT\\Base} & 64 & 0.6 & 110 & 1.0e-5 & 0.01 & 1000  & 128 \\
        \midrule
        \makecell{Amazon\\3M} & \makecell{BERT\\Base} & 128 & 0.35 & 50 & 5.0e-5 & 0.05 & 10000 & 128 \\
        \bottomrule
    \end{tabular}
    \label{hyperparameter_training}
\end{table}
We also present DST and other related settings in a separate Table \ref{dst_settings}. The learning rates for the auxiliary classifier and the intermediate layer are fixed at 5.01e-4 and 2.01e-4, respectively. We use a random growth mode with zero initialization, updating the topology until 66\% of the total training steps.

\begin{table}[ht]
    \centering
    \caption{DST and other related hyperparameter settings for different datasets.}
    \begin{tabular}{c|ccccc|cc|cc}
    \midrule
        Dataset & \makecell{Fan-in\\(sparsity)} & \makecell{Prune \\mode} & \makecell{Rewire\\threshold} & \makecell{Rewire\\fraction} & \makecell{Rewire \\interval}  & \makecell{Aux\\classifier} & \makecell{Aux\\cut-off} & \makecell{Inter.\\layer} & \makecell{Layer\\size}\\
        \midrule
        \makecell{Wiki10\\31K} & \makecell{64\\(0.92)} & fraction &  - & 0.25 & 300 & No & - & Yes & 4096 \\
        \midrule
        \makecell{LF\\AmazonTitles\\131K} & \makecell{128\\(0.83)} & fraction & - & 0.15 & 1000 & Yes & 12   & No & - \\
        \midrule
        \makecell{Wiki\\500K} & \makecell{128\\(0.83)} & threshold & 0.05 & - &  600 & Yes & 8  & No & - \\
        \midrule
        \makecell{Amazon\\670K} & \makecell{128\\(0.83)} & threshold & 0.01 & - &  800 & Yes & 15 & No  & - \\
        \midrule
        \makecell{Amazon\\3M} & \makecell{256\\(0.67)} & fraction & - & 0.25 & 1500 & Yes & 8 & No & - \\
        \bottomrule
    \end{tabular}
    \label{dst_settings}
\end{table}


\begin{figure}
  \centering
  \includegraphics[width=\textwidth]{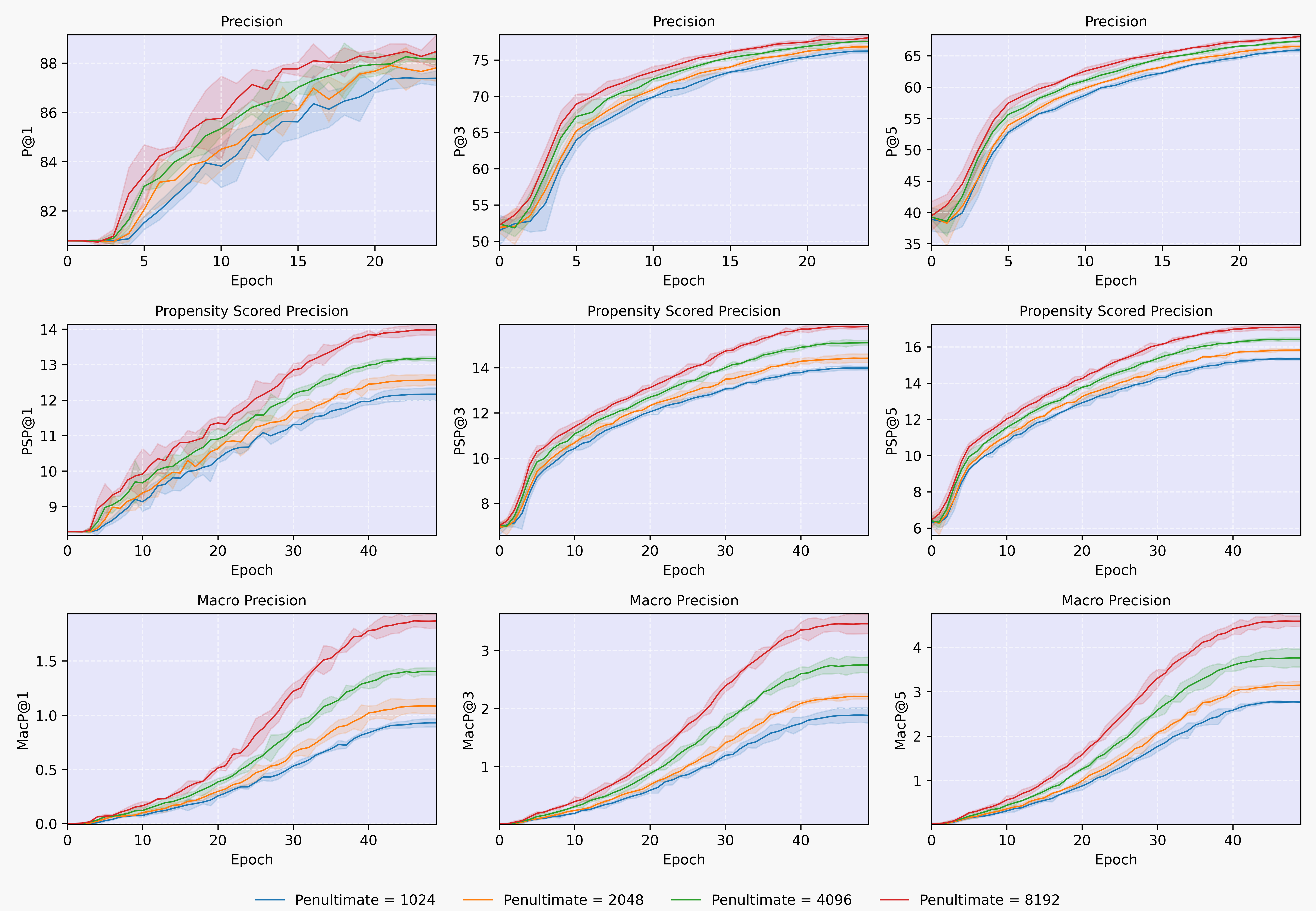}
  \caption{Impact of intermediate layer size on overall and tail label performance. The plots show precision, propensity-scored precision, and macro precision across epochs for different intermediate layer sizes (1024, 2048, 4096, and 8192).}
\label{intermediate_ablation}

\end{figure}

\section{Effect of Intermediate Layer Size on Overall and Tail Label Performance} We investigated the impact of varying sizes of the intermediate layer on both overall performance and the performance of tail labels specifically as shown in Figure \ref{intermediate_ablation}. It is important to highlight that although precision values peaked at certain layer sizes, the performance on tail labels continued to improve even beyond this point. This observation suggests that optimizing the size of the intermediate layer could play a crucial role in enhancing model effectiveness, particularly for tail labels which are often more challenging to predict accurately.


\section{The Role of Random Cluster based Meta Classifiers in XMC Problems}
\label{random_cluster} To understand the impact of random clusters on meta classifier-based methods, we selected the LightXML \cite{jiang2021lightxml} approach and experimented with two large-scale datasets: Amazon-670K and Wiki-500K. For our experiments, we used the original code from the official LightXML repository and the original clusters provided by the authors. We randomized the original clusters by applying several iterations of \texttt{random.shuffle()}, repeating the process twice to generate two sets of random clusters.

To ensure randomness, we calculated the intersection of elements between each pair of clusters from the original and random sets. We then took the maximum overlap value among all pairs, which was less than $3.5\%$ in both cases. Subsequently, we ran the LightXML code using the original clusters and the two sets of random clusters.

\begin{figure}
  \centering
  \includegraphics[width=\textwidth]{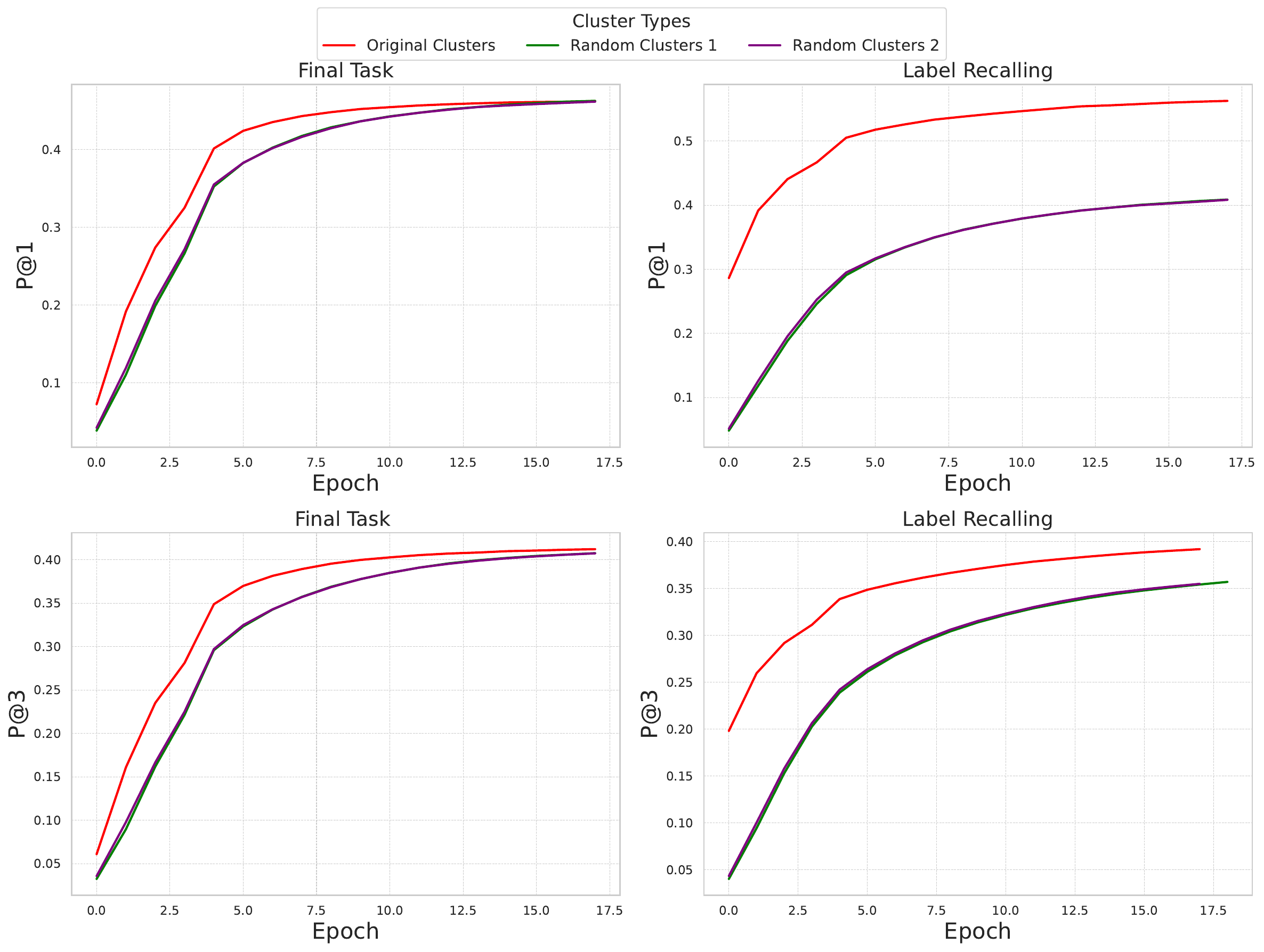}
  
  \caption{Final task and meta level precision performance for Amazon-670K}
  \label{cluster_ablation1}

\end{figure}

\begin{figure}
  \centering
  \includegraphics[width=\textwidth]{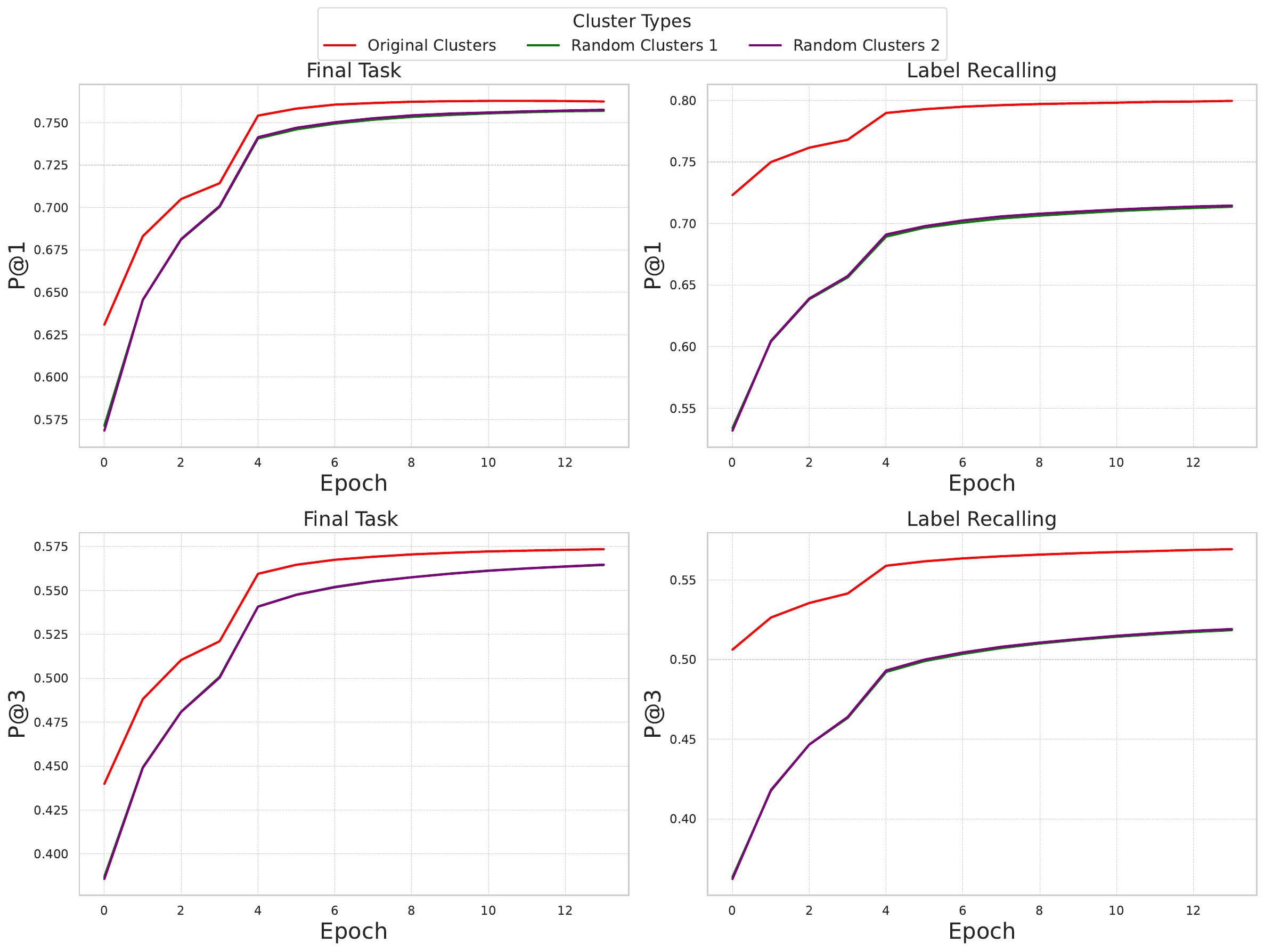}
  
  \caption{Final task and meta level precision performance for Wiki-500K}
  \label{cluster_ablation2}

\end{figure}

Our observations revealed that the final performance remained largely unaffected, although the learning process slowed down initially, as shown in the upper row of Figure \ref{cluster_ablation1}. The bottom row illustrates the precision of the meta classifier, which is lower for the random clusters as expected. We replicated the same experiment with the Wiki-500K dataset and observed similar results, which are also depicted in Figure \ref{cluster_ablation2}.

\section{Computational Resources}
While we want to demonstrate the memory efficiency of our algorithms, in order to enable meaningful comparison with existing methods,
we run all our experiments on a NVidia A100 GPU, and measure the memory consumption using \texttt{torch.cuda.max\_memory\_allocated}.
On this GPU, the experiments with Wiki31K take about 1 hour, Amazon-131K 8 hours, Amazon-670k 30 hours, Wikipedia-500k 36 hours and
Amazon-3M 72 hours.


\newpage
\section*{NeurIPS Paper Checklist}

The checklist is designed to encourage best practices for responsible machine learning research, addressing issues of reproducibility, transparency, research ethics, and societal impact. Do not remove the checklist: {\bf The papers not including the checklist will be desk rejected.} The checklist should follow the references and precede the (optional) supplemental material.  The checklist does NOT count towards the page
limit. 

Please read the checklist guidelines carefully for information on how to answer these questions. For each question in the checklist:
\begin{itemize}
    \item You should answer \answerYes{}, \answerNo{}, or \answerNA{}.
    \item \answerNA{} means either that the question is Not Applicable for that particular paper or the relevant information is Not Available.
    \item Please provide a short (1–2 sentence) justification right after your answer (even for NA). 
\end{itemize}

{\bf The checklist answers are an integral part of your paper submission.} They are visible to the reviewers, area chairs, senior area chairs, and ethics reviewers. You will be asked to also include it (after eventual revisions) with the final version of your paper, and its final version will be published with the paper.

The reviewers of your paper will be asked to use the checklist as one of the factors in their evaluation. While "\answerYes{}" is generally preferable to "\answerNo{}", it is perfectly acceptable to answer "\answerNo{}" provided a proper justification is given (e.g., "error bars are not reported because it would be too computationally expensive" or "we were unable to find the license for the dataset we used"). In general, answering "\answerNo{}" or "\answerNA{}" is not grounds for rejection. While the questions are phrased in a binary way, we acknowledge that the true answer is often more nuanced, so please just use your best judgment and write a justification to elaborate. All supporting evidence can appear either in the main paper or the supplemental material, provided in appendix. If you answer \answerYes{} to a question, in the justification please point to the section(s) where related material for the question can be found.

IMPORTANT, please:
\begin{itemize}
    \item {\bf Delete this instruction block, but keep the section heading ``NeurIPS paper checklist"},
    \item  {\bf Keep the checklist subsection headings, questions/answers and guidelines below.}
    \item {\bf Do not modify the questions and only use the provided macros for your answers}.
\end{itemize}


\begin{enumerate}

\item {\bf Claims}
    \item[] Question: Do the main claims made in the abstract and introduction accurately reflect the paper's contributions and scope?
    \item[] Answer: \answerYes{} 
    \item[] Justification: The abstract and introduction provide a suffcient explanation of the main idea of the paper. The problem setting and the contributions are explicitly stated in the introduction section of the paper.

\item {\bf Limitations}
    \item[] Question: Does the paper discuss the limitations of the work performed by the authors?
    \item[] Answer: \answerYes{} 
    \item[] Justification: A section discussing the limitations has been added towards the end of the paper. Various computational considerations are the main part of the paper, and are adequately discussed.

\item {\bf Theory Assumptions and Proofs}
    \item[] Question: For each theoretical result, does the paper provide the full set of assumptions and a complete (and correct) proof?
    \item[] Answer: \answerNA{} 
    \item[] Justification: There are no formal results proved or claimed in this paper

    \item {\bf Experimental Result Reproducibility}
    \item[] Question: Does the paper fully disclose all the information needed to reproduce the main experimental results of the paper to the extent that it affects the main claims and/or conclusions of the paper (regardless of whether the code and data are provided or not)?
    \item[] Answer: \answerYes{} 
    \item[] Justification: The dataset details, architecture outline and hyper-parameter details are sufficiently explained in the main body of the paper and appendices.

\item {\bf Open access to data and code}
    \item[] Question: Does the paper provide open access to the data and code, with sufficient instructions to faithfully reproduce the main experimental results, as described in supplemental material?
    \item[] Answer: Data-\answerYes{}, and code - \answerNo{}  
    \item[] Justification: We perform experiments on publicly available datasets. The code will be made publicly available in the near future.

\item {\bf Experimental Setting/Details}
    \item[] Question: Does the paper specify all the training and test details (e.g., data splits, hyperparameters, how they were chosen, type of optimizer, etc.) necessary to understand the results?
    \item[] Answer: \answerYes{} 
    \item[] Justification: The experimental setting and hyperparameter details are sufficiently explained in the main paper and appendices.

\item {\bf Experiment Statistical Significance}
    \item[] Question: Does the paper report error bars suitably and correctly defined or other appropriate information about the statistical significance of the experiments?
    \item[] Answer: \answerNo{} 
    \item[] Justification: While the train, and test splits are standard, due to the scale of datasets, and computations involved performing significance tests is not undertaken by the research community in this domain. 

\item {\bf Experiments Compute Resources}
    \item[] Question: For each experiment, does the paper provide sufficient information on the computer resources (type of compute workers, memory, time of execution) needed to reproduce the experiments?
    \item[] Answer: \answerYes{} 
    \item[] Justification: This has been sufficiently explained in the experiments section of the paper.
    
\item {\bf Code Of Ethics}
    \item[] Question: Does the research conducted in the paper conform, in every respect, with the NeurIPS Code of Ethics \url{https://neurips.cc/public/EthicsGuidelines}?
    \item[] Answer: \answerYes{} 
    \item[] Justification: Ethical considerations 

\item {\bf Broader Impacts}
    \item[] Question: Does the paper discuss both potential positive societal impacts and negative societal impacts of the work performed?
    \item[] Answer: \answerYes{} 
    \item[] Justification: A brief description of the societal impact of work in provided in the paper.
    
\item {\bf Safeguards}
    \item[] Question: Does the paper describe safeguards that have been put in place for responsible release of data or models that have a high risk for misuse (e.g., pretrained language models, image generators, or scraped datasets)?
    \item[] Answer: \answerNA{} 
    \item[] Justification: We do not release any Language model, or image generators.

\item {\bf Licenses for existing assets}
    \item[] Question: Are the creators or original owners of assets (e.g., code, data, models), used in the paper, properly credited and are the license and terms of use explicitly mentioned and properly respected?
    \item[] Answer: \answerYes{} 
    \item[] Justification: The link to the repository hosting publicly available datasets has been provided.

\item {\bf New Assets}
    \item[] Question: Are new assets introduced in the paper well documented and is the documentation provided alongside the assets?
    \item[] Answer: \answerNA{} 
    \item[] Justification: We do not release any new assests.

\item {\bf Crowdsourcing and Research with Human Subjects}
    \item[] Question: For crowdsourcing experiments and research with human subjects, does the paper include the full text of instructions given to participants and screenshots, if applicable, as well as details about compensation (if any)? 
    \item[] Answer: \answerNA{} 
    \item[] Justification: No study involving humans was part of this work.

\item {\bf Institutional Review Board (IRB) Approvals or Equivalent for Research with Human Subjects}
    \item[] Question: Does the paper describe potential risks incurred by study participants, whether such risks were disclosed to the subjects, and whether Institutional Review Board (IRB) approvals (or an equivalent approval/review based on the requirements of your country or institution) were obtained?
    \item[] Answer: \answerNA{} 
    \item[] Justification: No study involving humans was part of this work.

\end{enumerate}

\end{document}

%% file: figures/architectures.tex
\tikzset{input-node/.style={draw,rectangle, minimum height=0.5cm, minimum width=1.0cm}}
\tikzset{weight-node/.style={draw,rectangle, minimum height=0.5cm, minimum width=1.0cm}}
\tikzset{decode-node/.style={draw,trapezium, shape border rotate=0, minimum height=0.5cm, inner sep=0pt}}
\tikzset{dense-style/.style={fill=gray!50!white}}
\tikzset{sparse-style/.style={pattern=dots, pattern color=gray}}
\tikzset{node distance=1.5cm}
\begin{tikzpicture}[->, semithick, node distance=0.9cm, font=\small]
    \node[weight-node](w) {encoder};
    \node[decode-node, below of=w, minimum width=1.25cm, trapezium angle=60, trapezium stretches=true, dense-style](w1) {$W$};
    \node[decode-node, sparse-style, below=1.5cm of w, minimum width=2.0cm, trapezium angle=70, trapezium stretches=true](out) {\contour{white}{$S$}};
    \draw (w) -- (w1);
    \draw (w1) -- (out);
    \node[above=0.5cm of w.center, minimum height=0.5cm](lbl1) {$+{}$Intermediate};

    \node[weight-node, left=1.1cm of w](w2) {encoder};
    \node[decode-node, sparse-style, below=1.5cm of w2, minimum width=1.8cm, trapezium angle=60, trapezium stretches=true](out2) {\contour{white}{$S$}};
    \draw (w2) -- (out2);
    \node[above=0.5cm of w2.center, minimum height=0.5cm](lbl2) {DST};

    \node[weight-node, right=1.1cm of w](w3) {encoder};
    \node[decode-node, sparse-style, below=1.5cm of w3, minimum width=1.8cm, trapezium angle=60, trapezium stretches=true](out3) {\contour{white}{$S$}};
    \node[decode-node, dense-style, right=0.45cm of out3, yshift=0.33cm, minimum width=1.0cm, trapezium angle=60, trapezium stretches=true](aux) {Aux};
    \draw (w3) -- (out3);
    \node[above=0.5cm of w3.center, minimum height=0.5cm](lbl3) {$+{}$Auxiliary};
    \node[below=0.75cm of w3.center, fill,circle, inner sep=1pt](mid3) {};
    \draw (mid3) --  (mid3-|aux) -- (aux);

    \draw (-3.3,-3) -- (4.3, -3);
    \node[below=2.8cm of w]() {label space size};
    \node[below=2.25cm of w]() {\footnotesize{31k}};
    \node[below=2.25cm of w2]() {\footnotesize{4k}};
    \node[below=2.25cm of w3]() {\footnotesize{670k}};
    \draw[-,thin] (-1.2, 0.8) -- (-1.2, -2.8);
    \draw[-,thin] (1.2, 0.8) -- (1.2, -2.8);
\end{tikzpicture}

%% file: figures/intro_plot.pgf
\begingroup%
\makeatletter%
\begin{pgfpicture}%
\pgfpathrectangle{\pgfpointorigin}{\pgfqpoint{6.000000in}{5.000000in}}%
\pgfusepath{use as bounding box, clip}%
\begin{pgfscope}%
\pgfsetbuttcap%
\pgfsetmiterjoin%
\definecolor{currentfill}{rgb}{1.000000,1.000000,1.000000}%
\pgfsetfillcolor{currentfill}%
\pgfsetlinewidth{0.000000pt}%
\definecolor{currentstroke}{rgb}{1.000000,1.000000,1.000000}%
\pgfsetstrokecolor{currentstroke}%
\pgfsetdash{}{0pt}%
\pgfpathmoveto{\pgfqpoint{0.000000in}{0.000000in}}%
\pgfpathlineto{\pgfqpoint{6.000000in}{0.000000in}}%
\pgfpathlineto{\pgfqpoint{6.000000in}{5.000000in}}%
\pgfpathlineto{\pgfqpoint{0.000000in}{5.000000in}}%
\pgfpathlineto{\pgfqpoint{0.000000in}{0.000000in}}%
\pgfpathclose%
\pgfusepath{fill}%
\end{pgfscope}%
\begin{pgfscope}%
\pgfsetbuttcap%
\pgfsetmiterjoin%
\definecolor{currentfill}{rgb}{0.917647,0.917647,0.949020}%
\pgfsetfillcolor{currentfill}%
\pgfsetlinewidth{0.000000pt}%
\definecolor{currentstroke}{rgb}{0.000000,0.000000,0.000000}%
\pgfsetstrokecolor{currentstroke}%
\pgfsetstrokeopacity{0.000000}%
\pgfsetdash{}{0pt}%
\pgfpathmoveto{\pgfqpoint{0.750000in}{0.550000in}}%
\pgfpathlineto{\pgfqpoint{5.400000in}{0.550000in}}%
\pgfpathlineto{\pgfqpoint{5.400000in}{4.400000in}}%
\pgfpathlineto{\pgfqpoint{0.750000in}{4.400000in}}%
\pgfpathlineto{\pgfqpoint{0.750000in}{0.550000in}}%
\pgfpathclose%
\pgfusepath{fill}%
\end{pgfscope}%
\begin{pgfscope}%
\pgfpathrectangle{\pgfqpoint{0.750000in}{0.550000in}}{\pgfqpoint{4.650000in}{3.850000in}}%
\pgfusepath{clip}%
\pgfsetroundcap%
\pgfsetroundjoin%
\pgfsetlinewidth{1.003750pt}%
\definecolor{currentstroke}{rgb}{1.000000,1.000000,1.000000}%
\pgfsetstrokecolor{currentstroke}%
\pgfsetdash{}{0pt}%
\pgfpathmoveto{\pgfqpoint{1.398668in}{0.550000in}}%
\pgfpathlineto{\pgfqpoint{1.398668in}{4.400000in}}%
\pgfusepath{stroke}%
\end{pgfscope}%
\begin{pgfscope}%
\definecolor{textcolor}{rgb}{0.150000,0.150000,0.150000}%
\pgfsetstrokecolor{textcolor}%
\pgfsetfillcolor{textcolor}%
\pgftext[x=1.398668in,y=0.418056in,,top]{\color{textcolor}{\sffamily\fontsize{11.000000}{13.200000}\selectfont\catcode`\^=\active\def^{\ifmmode\sp\else\^{}\fi}\catcode`\%=\active\def
\end{pgfscope}%
\begin{pgfscope}%
\pgfpathrectangle{\pgfqpoint{0.750000in}{0.550000in}}{\pgfqpoint{4.650000in}{3.850000in}}%
\pgfusepath{clip}%
\pgfsetroundcap%
\pgfsetroundjoin%
\pgfsetlinewidth{1.003750pt}%
\definecolor{currentstroke}{rgb}{1.000000,1.000000,1.000000}%
\pgfsetstrokecolor{currentstroke}%
\pgfsetdash{}{0pt}%
\pgfpathmoveto{\pgfqpoint{2.127508in}{0.550000in}}%
\pgfpathlineto{\pgfqpoint{2.127508in}{4.400000in}}%
\pgfusepath{stroke}%
\end{pgfscope}%
\begin{pgfscope}%
\definecolor{textcolor}{rgb}{0.150000,0.150000,0.150000}%
\pgfsetstrokecolor{textcolor}%
\pgfsetfillcolor{textcolor}%
\pgftext[x=2.127508in,y=0.418056in,,top]{\color{textcolor}{\sffamily\fontsize{11.000000}{13.200000}\selectfont\catcode`\^=\active\def^{\ifmmode\sp\else\^{}\fi}\catcode`\%=\active\def
\end{pgfscope}%
\begin{pgfscope}%
\pgfpathrectangle{\pgfqpoint{0.750000in}{0.550000in}}{\pgfqpoint{4.650000in}{3.850000in}}%
\pgfusepath{clip}%
\pgfsetroundcap%
\pgfsetroundjoin%
\pgfsetlinewidth{1.003750pt}%
\definecolor{currentstroke}{rgb}{1.000000,1.000000,1.000000}%
\pgfsetstrokecolor{currentstroke}%
\pgfsetdash{}{0pt}%
\pgfpathmoveto{\pgfqpoint{2.856348in}{0.550000in}}%
\pgfpathlineto{\pgfqpoint{2.856348in}{4.400000in}}%
\pgfusepath{stroke}%
\end{pgfscope}%
\begin{pgfscope}%
\definecolor{textcolor}{rgb}{0.150000,0.150000,0.150000}%
\pgfsetstrokecolor{textcolor}%
\pgfsetfillcolor{textcolor}%
\pgftext[x=2.856348in,y=0.418056in,,top]{\color{textcolor}{\sffamily\fontsize{11.000000}{13.200000}\selectfont\catcode`\^=\active\def^{\ifmmode\sp\else\^{}\fi}\catcode`\%=\active\def
\end{pgfscope}%
\begin{pgfscope}%
\pgfpathrectangle{\pgfqpoint{0.750000in}{0.550000in}}{\pgfqpoint{4.650000in}{3.850000in}}%
\pgfusepath{clip}%
\pgfsetroundcap%
\pgfsetroundjoin%
\pgfsetlinewidth{1.003750pt}%
\definecolor{currentstroke}{rgb}{1.000000,1.000000,1.000000}%
\pgfsetstrokecolor{currentstroke}%
\pgfsetdash{}{0pt}%
\pgfpathmoveto{\pgfqpoint{3.585188in}{0.550000in}}%
\pgfpathlineto{\pgfqpoint{3.585188in}{4.400000in}}%
\pgfusepath{stroke}%
\end{pgfscope}%
\begin{pgfscope}%
\definecolor{textcolor}{rgb}{0.150000,0.150000,0.150000}%
\pgfsetstrokecolor{textcolor}%
\pgfsetfillcolor{textcolor}%
\pgftext[x=3.585188in,y=0.418056in,,top]{\color{textcolor}{\sffamily\fontsize{11.000000}{13.200000}\selectfont\catcode`\^=\active\def^{\ifmmode\sp\else\^{}\fi}\catcode`\%=\active\def
\end{pgfscope}%
\begin{pgfscope}%
\pgfpathrectangle{\pgfqpoint{0.750000in}{0.550000in}}{\pgfqpoint{4.650000in}{3.850000in}}%
\pgfusepath{clip}%
\pgfsetroundcap%
\pgfsetroundjoin%
\pgfsetlinewidth{1.003750pt}%
\definecolor{currentstroke}{rgb}{1.000000,1.000000,1.000000}%
\pgfsetstrokecolor{currentstroke}%
\pgfsetdash{}{0pt}%
\pgfpathmoveto{\pgfqpoint{4.314028in}{0.550000in}}%
\pgfpathlineto{\pgfqpoint{4.314028in}{4.400000in}}%
\pgfusepath{stroke}%
\end{pgfscope}%
\begin{pgfscope}%
\definecolor{textcolor}{rgb}{0.150000,0.150000,0.150000}%
\pgfsetstrokecolor{textcolor}%
\pgfsetfillcolor{textcolor}%
\pgftext[x=4.314028in,y=0.418056in,,top]{\color{textcolor}{\sffamily\fontsize{11.000000}{13.200000}\selectfont\catcode`\^=\active\def^{\ifmmode\sp\else\^{}\fi}\catcode`\%=\active\def
\end{pgfscope}%
\begin{pgfscope}%
\pgfpathrectangle{\pgfqpoint{0.750000in}{0.550000in}}{\pgfqpoint{4.650000in}{3.850000in}}%
\pgfusepath{clip}%
\pgfsetroundcap%
\pgfsetroundjoin%
\pgfsetlinewidth{1.003750pt}%
\definecolor{currentstroke}{rgb}{1.000000,1.000000,1.000000}%
\pgfsetstrokecolor{currentstroke}%
\pgfsetdash{}{0pt}%
\pgfpathmoveto{\pgfqpoint{5.042868in}{0.550000in}}%
\pgfpathlineto{\pgfqpoint{5.042868in}{4.400000in}}%
\pgfusepath{stroke}%
\end{pgfscope}%
\begin{pgfscope}%
\definecolor{textcolor}{rgb}{0.150000,0.150000,0.150000}%
\pgfsetstrokecolor{textcolor}%
\pgfsetfillcolor{textcolor}%
\pgftext[x=5.042868in,y=0.418056in,,top]{\color{textcolor}{\sffamily\fontsize{11.000000}{13.200000}\selectfont\catcode`\^=\active\def^{\ifmmode\sp\else\^{}\fi}\catcode`\%=\active\def
\end{pgfscope}%
\begin{pgfscope}%
\definecolor{textcolor}{rgb}{0.150000,0.150000,0.150000}%
\pgfsetstrokecolor{textcolor}%
\pgfsetfillcolor{textcolor}%
\pgftext[x=3.075000in,y=0.222777in,,top]{\color{textcolor}{\sffamily\fontsize{16.000000}{19.200000}\selectfont\catcode`\^=\active\def^{\ifmmode\sp\else\^{}\fi}\catcode`\%=\active\def
\end{pgfscope}%
\begin{pgfscope}%
\pgfpathrectangle{\pgfqpoint{0.750000in}{0.550000in}}{\pgfqpoint{4.650000in}{3.850000in}}%
\pgfusepath{clip}%
\pgfsetroundcap%
\pgfsetroundjoin%
\pgfsetlinewidth{1.003750pt}%
\definecolor{currentstroke}{rgb}{1.000000,1.000000,1.000000}%
\pgfsetstrokecolor{currentstroke}%
\pgfsetdash{}{0pt}%
\pgfpathmoveto{\pgfqpoint{0.750000in}{1.151829in}}%
\pgfpathlineto{\pgfqpoint{5.400000in}{1.151829in}}%
\pgfusepath{stroke}%
\end{pgfscope}%
\begin{pgfscope}%
\definecolor{textcolor}{rgb}{0.150000,0.150000,0.150000}%
\pgfsetstrokecolor{textcolor}%
\pgfsetfillcolor{textcolor}%
\pgftext[x=0.405674in, y=1.097149in, left, base]{\color{textcolor}{\sffamily\fontsize{11.000000}{13.200000}\selectfont\catcode`\^=\active\def^{\ifmmode\sp\else\^{}\fi}\catcode`\%=\active\def
\end{pgfscope}%
\begin{pgfscope}%
\pgfpathrectangle{\pgfqpoint{0.750000in}{0.550000in}}{\pgfqpoint{4.650000in}{3.850000in}}%
\pgfusepath{clip}%
\pgfsetroundcap%
\pgfsetroundjoin%
\pgfsetlinewidth{1.003750pt}%
\definecolor{currentstroke}{rgb}{1.000000,1.000000,1.000000}%
\pgfsetstrokecolor{currentstroke}%
\pgfsetdash{}{0pt}%
\pgfpathmoveto{\pgfqpoint{0.750000in}{1.927882in}}%
\pgfpathlineto{\pgfqpoint{5.400000in}{1.927882in}}%
\pgfusepath{stroke}%
\end{pgfscope}%
\begin{pgfscope}%
\definecolor{textcolor}{rgb}{0.150000,0.150000,0.150000}%
\pgfsetstrokecolor{textcolor}%
\pgfsetfillcolor{textcolor}%
\pgftext[x=0.405674in, y=1.873202in, left, base]{\color{textcolor}{\sffamily\fontsize{11.000000}{13.200000}\selectfont\catcode`\^=\active\def^{\ifmmode\sp\else\^{}\fi}\catcode`\%=\active\def
\end{pgfscope}%
\begin{pgfscope}%
\pgfpathrectangle{\pgfqpoint{0.750000in}{0.550000in}}{\pgfqpoint{4.650000in}{3.850000in}}%
\pgfusepath{clip}%
\pgfsetroundcap%
\pgfsetroundjoin%
\pgfsetlinewidth{1.003750pt}%
\definecolor{currentstroke}{rgb}{1.000000,1.000000,1.000000}%
\pgfsetstrokecolor{currentstroke}%
\pgfsetdash{}{0pt}%
\pgfpathmoveto{\pgfqpoint{0.750000in}{2.703936in}}%
\pgfpathlineto{\pgfqpoint{5.400000in}{2.703936in}}%
\pgfusepath{stroke}%
\end{pgfscope}%
\begin{pgfscope}%
\definecolor{textcolor}{rgb}{0.150000,0.150000,0.150000}%
\pgfsetstrokecolor{textcolor}%
\pgfsetfillcolor{textcolor}%
\pgftext[x=0.405674in, y=2.649255in, left, base]{\color{textcolor}{\sffamily\fontsize{11.000000}{13.200000}\selectfont\catcode`\^=\active\def^{\ifmmode\sp\else\^{}\fi}\catcode`\%=\active\def
\end{pgfscope}%
\begin{pgfscope}%
\pgfpathrectangle{\pgfqpoint{0.750000in}{0.550000in}}{\pgfqpoint{4.650000in}{3.850000in}}%
\pgfusepath{clip}%
\pgfsetroundcap%
\pgfsetroundjoin%
\pgfsetlinewidth{1.003750pt}%
\definecolor{currentstroke}{rgb}{1.000000,1.000000,1.000000}%
\pgfsetstrokecolor{currentstroke}%
\pgfsetdash{}{0pt}%
\pgfpathmoveto{\pgfqpoint{0.750000in}{3.479989in}}%
\pgfpathlineto{\pgfqpoint{5.400000in}{3.479989in}}%
\pgfusepath{stroke}%
\end{pgfscope}%
\begin{pgfscope}%
\definecolor{textcolor}{rgb}{0.150000,0.150000,0.150000}%
\pgfsetstrokecolor{textcolor}%
\pgfsetfillcolor{textcolor}%
\pgftext[x=0.405674in, y=3.425308in, left, base]{\color{textcolor}{\sffamily\fontsize{11.000000}{13.200000}\selectfont\catcode`\^=\active\def^{\ifmmode\sp\else\^{}\fi}\catcode`\%=\active\def
\end{pgfscope}%
\begin{pgfscope}%
\pgfpathrectangle{\pgfqpoint{0.750000in}{0.550000in}}{\pgfqpoint{4.650000in}{3.850000in}}%
\pgfusepath{clip}%
\pgfsetroundcap%
\pgfsetroundjoin%
\pgfsetlinewidth{1.003750pt}%
\definecolor{currentstroke}{rgb}{1.000000,1.000000,1.000000}%
\pgfsetstrokecolor{currentstroke}%
\pgfsetdash{}{0pt}%
\pgfpathmoveto{\pgfqpoint{0.750000in}{4.256042in}}%
\pgfpathlineto{\pgfqpoint{5.400000in}{4.256042in}}%
\pgfusepath{stroke}%
\end{pgfscope}%
\begin{pgfscope}%
\definecolor{textcolor}{rgb}{0.150000,0.150000,0.150000}%
\pgfsetstrokecolor{textcolor}%
\pgfsetfillcolor{textcolor}%
\pgftext[x=0.405674in, y=4.201361in, left, base]{\color{textcolor}{\sffamily\fontsize{11.000000}{13.200000}\selectfont\catcode`\^=\active\def^{\ifmmode\sp\else\^{}\fi}\catcode`\%=\active\def
\end{pgfscope}%
\begin{pgfscope}%
\definecolor{textcolor}{rgb}{0.150000,0.150000,0.150000}%
\pgfsetstrokecolor{textcolor}%
\pgfsetfillcolor{textcolor}%
\pgftext[x=0.350118in,y=2.475000in,,bottom,rotate=90.000000]{\color{textcolor}{\sffamily\fontsize{16.000000}{19.200000}\selectfont\catcode`\^=\active\def^{\ifmmode\sp\else\^{}\fi}\catcode`\%=\active\def
\end{pgfscope}%
\begin{pgfscope}%
\pgfpathrectangle{\pgfqpoint{0.750000in}{0.550000in}}{\pgfqpoint{4.650000in}{3.850000in}}%
\pgfusepath{clip}%
\pgfsetbuttcap%
\pgfsetroundjoin%
\pgfsetlinewidth{1.505625pt}%
\definecolor{currentstroke}{rgb}{0.000000,0.000000,0.000000}%
\pgfsetstrokecolor{currentstroke}%
\pgfsetdash{{5.550000pt}{2.400000pt}}{0.000000pt}%
\pgfpathmoveto{\pgfqpoint{0.961364in}{4.225000in}}%
\pgfpathlineto{\pgfqpoint{3.293652in}{4.225000in}}%
\pgfpathlineto{\pgfqpoint{4.605564in}{4.225000in}}%
\pgfpathlineto{\pgfqpoint{5.188636in}{4.225000in}}%
\pgfusepath{stroke}%
\end{pgfscope}%
\begin{pgfscope}%
\pgfpathrectangle{\pgfqpoint{0.750000in}{0.550000in}}{\pgfqpoint{4.650000in}{3.850000in}}%
\pgfusepath{clip}%
\pgfsetroundcap%
\pgfsetroundjoin%
\pgfsetlinewidth{1.505625pt}%
\definecolor{currentstroke}{rgb}{0.501961,0.000000,0.501961}%
\pgfsetstrokecolor{currentstroke}%
\pgfsetdash{}{0pt}%
\pgfpathmoveto{\pgfqpoint{0.961364in}{3.906818in}}%
\pgfpathlineto{\pgfqpoint{3.293652in}{3.239412in}}%
\pgfpathlineto{\pgfqpoint{4.605564in}{1.392406in}}%
\pgfpathlineto{\pgfqpoint{5.188636in}{0.725000in}}%
\pgfusepath{stroke}%
\end{pgfscope}%
\begin{pgfscope}%
\pgfpathrectangle{\pgfqpoint{0.750000in}{0.550000in}}{\pgfqpoint{4.650000in}{3.850000in}}%
\pgfusepath{clip}%
\pgfsetbuttcap%
\pgfsetbeveljoin%
\definecolor{currentfill}{rgb}{0.501961,0.000000,0.501961}%
\pgfsetfillcolor{currentfill}%
\pgfsetlinewidth{1.003750pt}%
\definecolor{currentstroke}{rgb}{0.501961,0.000000,0.501961}%
\pgfsetstrokecolor{currentstroke}%
\pgfsetdash{}{0pt}%
\pgfsys@defobject{currentmarker}{\pgfqpoint{-0.039627in}{-0.033709in}}{\pgfqpoint{0.039627in}{0.041667in}}{%
\pgfpathmoveto{\pgfqpoint{0.000000in}{0.041667in}}%
\pgfpathlineto{\pgfqpoint{-0.009355in}{0.012876in}}%
\pgfpathlineto{\pgfqpoint{-0.039627in}{0.012876in}}%
\pgfpathlineto{\pgfqpoint{-0.015136in}{-0.004918in}}%
\pgfpathlineto{\pgfqpoint{-0.024491in}{-0.033709in}}%
\pgfpathlineto{\pgfqpoint{-0.000000in}{-0.015915in}}%
\pgfpathlineto{\pgfqpoint{0.024491in}{-0.033709in}}%
\pgfpathlineto{\pgfqpoint{0.015136in}{-0.004918in}}%
\pgfpathlineto{\pgfqpoint{0.039627in}{0.012876in}}%
\pgfpathlineto{\pgfqpoint{0.009355in}{0.012876in}}%
\pgfpathlineto{\pgfqpoint{0.000000in}{0.041667in}}%
\pgfpathclose%
\pgfusepath{stroke,fill}%
}%
\begin{pgfscope}%
\pgfsys@transformshift{0.961364in}{3.906818in}%
\pgfsys@useobject{currentmarker}{}%
\end{pgfscope}%
\begin{pgfscope}%
\pgfsys@transformshift{3.293652in}{3.239412in}%
\pgfsys@useobject{currentmarker}{}%
\end{pgfscope}%
\begin{pgfscope}%
\pgfsys@transformshift{4.605564in}{1.392406in}%
\pgfsys@useobject{currentmarker}{}%
\end{pgfscope}%
\begin{pgfscope}%
\pgfsys@transformshift{5.188636in}{0.725000in}%
\pgfsys@useobject{currentmarker}{}%
\end{pgfscope}%
\end{pgfscope}%
\begin{pgfscope}%
\pgfpathrectangle{\pgfqpoint{0.750000in}{0.550000in}}{\pgfqpoint{4.650000in}{3.850000in}}%
\pgfusepath{clip}%
\pgfsetroundcap%
\pgfsetroundjoin%
\pgfsetlinewidth{1.505625pt}%
\definecolor{currentstroke}{rgb}{1.000000,0.000000,0.000000}%
\pgfsetstrokecolor{currentstroke}%
\pgfsetdash{}{0pt}%
\pgfpathmoveto{\pgfqpoint{0.961364in}{4.069789in}}%
\pgfpathlineto{\pgfqpoint{3.293652in}{4.038747in}}%
\pgfpathlineto{\pgfqpoint{4.605564in}{3.790410in}}%
\pgfpathlineto{\pgfqpoint{5.188636in}{3.324778in}}%
\pgfusepath{stroke}%
\end{pgfscope}%
\begin{pgfscope}%
\pgfpathrectangle{\pgfqpoint{0.750000in}{0.550000in}}{\pgfqpoint{4.650000in}{3.850000in}}%
\pgfusepath{clip}%
\pgfsetbuttcap%
\pgfsetmiterjoin%
\definecolor{currentfill}{rgb}{1.000000,0.000000,0.000000}%
\pgfsetfillcolor{currentfill}%
\pgfsetlinewidth{1.003750pt}%
\definecolor{currentstroke}{rgb}{1.000000,0.000000,0.000000}%
\pgfsetstrokecolor{currentstroke}%
\pgfsetdash{}{0pt}%
\pgfsys@defobject{currentmarker}{\pgfqpoint{-0.058926in}{-0.058926in}}{\pgfqpoint{0.058926in}{0.058926in}}{%
\pgfpathmoveto{\pgfqpoint{-0.000000in}{-0.058926in}}%
\pgfpathlineto{\pgfqpoint{0.058926in}{0.000000in}}%
\pgfpathlineto{\pgfqpoint{0.000000in}{0.058926in}}%
\pgfpathlineto{\pgfqpoint{-0.058926in}{0.000000in}}%
\pgfpathlineto{\pgfqpoint{-0.000000in}{-0.058926in}}%
\pgfpathclose%
\pgfusepath{stroke,fill}%
}%
\begin{pgfscope}%
\pgfsys@transformshift{0.961364in}{4.069789in}%
\pgfsys@useobject{currentmarker}{}%
\end{pgfscope}%
\begin{pgfscope}%
\pgfsys@transformshift{3.293652in}{4.038747in}%
\pgfsys@useobject{currentmarker}{}%
\end{pgfscope}%
\begin{pgfscope}%
\pgfsys@transformshift{4.605564in}{3.790410in}%
\pgfsys@useobject{currentmarker}{}%
\end{pgfscope}%
\begin{pgfscope}%
\pgfsys@transformshift{5.188636in}{3.324778in}%
\pgfsys@useobject{currentmarker}{}%
\end{pgfscope}%
\end{pgfscope}%
\begin{pgfscope}%
\pgfpathrectangle{\pgfqpoint{0.750000in}{0.550000in}}{\pgfqpoint{4.650000in}{3.850000in}}%
\pgfusepath{clip}%
\pgfsetroundcap%
\pgfsetroundjoin%
\pgfsetlinewidth{1.505625pt}%
\definecolor{currentstroke}{rgb}{1.000000,0.647059,0.000000}%
\pgfsetstrokecolor{currentstroke}%
\pgfsetdash{}{0pt}%
\pgfpathmoveto{\pgfqpoint{0.961364in}{4.023226in}}%
\pgfpathlineto{\pgfqpoint{3.293652in}{3.914579in}}%
\pgfpathlineto{\pgfqpoint{4.605564in}{3.596397in}}%
\pgfpathlineto{\pgfqpoint{5.188636in}{0.802605in}}%
\pgfusepath{stroke}%
\end{pgfscope}%
\begin{pgfscope}%
\pgfpathrectangle{\pgfqpoint{0.750000in}{0.550000in}}{\pgfqpoint{4.650000in}{3.850000in}}%
\pgfusepath{clip}%
\pgfsetbuttcap%
\pgfsetroundjoin%
\definecolor{currentfill}{rgb}{1.000000,0.647059,0.000000}%
\pgfsetfillcolor{currentfill}%
\pgfsetlinewidth{1.003750pt}%
\definecolor{currentstroke}{rgb}{1.000000,0.647059,0.000000}%
\pgfsetstrokecolor{currentstroke}%
\pgfsetdash{}{0pt}%
\pgfsys@defobject{currentmarker}{\pgfqpoint{-0.041667in}{-0.041667in}}{\pgfqpoint{0.041667in}{0.041667in}}{%
\pgfpathmoveto{\pgfqpoint{0.000000in}{-0.041667in}}%
\pgfpathcurveto{\pgfqpoint{0.011050in}{-0.041667in}}{\pgfqpoint{0.021649in}{-0.037276in}}{\pgfqpoint{0.029463in}{-0.029463in}}%
\pgfpathcurveto{\pgfqpoint{0.037276in}{-0.021649in}}{\pgfqpoint{0.041667in}{-0.011050in}}{\pgfqpoint{0.041667in}{0.000000in}}%
\pgfpathcurveto{\pgfqpoint{0.041667in}{0.011050in}}{\pgfqpoint{0.037276in}{0.021649in}}{\pgfqpoint{0.029463in}{0.029463in}}%
\pgfpathcurveto{\pgfqpoint{0.021649in}{0.037276in}}{\pgfqpoint{0.011050in}{0.041667in}}{\pgfqpoint{0.000000in}{0.041667in}}%
\pgfpathcurveto{\pgfqpoint{-0.011050in}{0.041667in}}{\pgfqpoint{-0.021649in}{0.037276in}}{\pgfqpoint{-0.029463in}{0.029463in}}%
\pgfpathcurveto{\pgfqpoint{-0.037276in}{0.021649in}}{\pgfqpoint{-0.041667in}{0.011050in}}{\pgfqpoint{-0.041667in}{0.000000in}}%
\pgfpathcurveto{\pgfqpoint{-0.041667in}{-0.011050in}}{\pgfqpoint{-0.037276in}{-0.021649in}}{\pgfqpoint{-0.029463in}{-0.029463in}}%
\pgfpathcurveto{\pgfqpoint{-0.021649in}{-0.037276in}}{\pgfqpoint{-0.011050in}{-0.041667in}}{\pgfqpoint{0.000000in}{-0.041667in}}%
\pgfpathlineto{\pgfqpoint{0.000000in}{-0.041667in}}%
\pgfpathclose%
\pgfusepath{stroke,fill}%
}%
\begin{pgfscope}%
\pgfsys@transformshift{0.961364in}{4.023226in}%
\pgfsys@useobject{currentmarker}{}%
\end{pgfscope}%
\begin{pgfscope}%
\pgfsys@transformshift{3.293652in}{3.914579in}%
\pgfsys@useobject{currentmarker}{}%
\end{pgfscope}%
\begin{pgfscope}%
\pgfsys@transformshift{4.605564in}{3.596397in}%
\pgfsys@useobject{currentmarker}{}%
\end{pgfscope}%
\begin{pgfscope}%
\pgfsys@transformshift{5.188636in}{0.802605in}%
\pgfsys@useobject{currentmarker}{}%
\end{pgfscope}%
\end{pgfscope}%
\begin{pgfscope}%
\pgfsetrectcap%
\pgfsetmiterjoin%
\pgfsetlinewidth{1.254687pt}%
\definecolor{currentstroke}{rgb}{1.000000,1.000000,1.000000}%
\pgfsetstrokecolor{currentstroke}%
\pgfsetdash{}{0pt}%
\pgfpathmoveto{\pgfqpoint{0.750000in}{0.550000in}}%
\pgfpathlineto{\pgfqpoint{0.750000in}{4.400000in}}%
\pgfusepath{stroke}%
\end{pgfscope}%
\begin{pgfscope}%
\pgfsetrectcap%
\pgfsetmiterjoin%
\pgfsetlinewidth{1.254687pt}%
\definecolor{currentstroke}{rgb}{1.000000,1.000000,1.000000}%
\pgfsetstrokecolor{currentstroke}%
\pgfsetdash{}{0pt}%
\pgfpathmoveto{\pgfqpoint{5.400000in}{0.550000in}}%
\pgfpathlineto{\pgfqpoint{5.400000in}{4.400000in}}%
\pgfusepath{stroke}%
\end{pgfscope}%
\begin{pgfscope}%
\pgfsetrectcap%
\pgfsetmiterjoin%
\pgfsetlinewidth{1.254687pt}%
\definecolor{currentstroke}{rgb}{1.000000,1.000000,1.000000}%
\pgfsetstrokecolor{currentstroke}%
\pgfsetdash{}{0pt}%
\pgfpathmoveto{\pgfqpoint{0.750000in}{0.550000in}}%
\pgfpathlineto{\pgfqpoint{5.400000in}{0.550000in}}%
\pgfusepath{stroke}%
\end{pgfscope}%
\begin{pgfscope}%
\pgfsetrectcap%
\pgfsetmiterjoin%
\pgfsetlinewidth{1.254687pt}%
\definecolor{currentstroke}{rgb}{1.000000,1.000000,1.000000}%
\pgfsetstrokecolor{currentstroke}%
\pgfsetdash{}{0pt}%
\pgfpathmoveto{\pgfqpoint{0.750000in}{4.400000in}}%
\pgfpathlineto{\pgfqpoint{5.400000in}{4.400000in}}%
\pgfusepath{stroke}%
\end{pgfscope}%
\begin{pgfscope}%
\definecolor{textcolor}{rgb}{0.150000,0.150000,0.150000}%
\pgfsetstrokecolor{textcolor}%
\pgfsetfillcolor{textcolor}%
\pgftext[x=3.075000in,y=4.483333in,,base]{\color{textcolor}{\sffamily\fontsize{16.000000}{19.200000}\selectfont\catcode`\^=\active\def^{\ifmmode\sp\else\^{}\fi}\catcode`\%=\active\def
\end{pgfscope}%
\begin{pgfscope}%
\pgfsetbuttcap%
\pgfsetmiterjoin%
\definecolor{currentfill}{rgb}{0.917647,0.917647,0.949020}%
\pgfsetfillcolor{currentfill}%
\pgfsetfillopacity{0.800000}%
\pgfsetlinewidth{1.003750pt}%
\definecolor{currentstroke}{rgb}{0.800000,0.800000,0.800000}%
\pgfsetstrokecolor{currentstroke}%
\pgfsetstrokeopacity{0.800000}%
\pgfsetdash{}{0pt}%
\pgfpathmoveto{\pgfqpoint{0.890000in}{0.650000in}}%
\pgfpathlineto{\pgfqpoint{2.808222in}{0.650000in}}%
\pgfpathquadraticcurveto{\pgfqpoint{2.848222in}{0.650000in}}{\pgfqpoint{2.848222in}{0.690000in}}%
\pgfpathlineto{\pgfqpoint{2.848222in}{1.811211in}}%
\pgfpathquadraticcurveto{\pgfqpoint{2.848222in}{1.851211in}}{\pgfqpoint{2.808222in}{1.851211in}}%
\pgfpathlineto{\pgfqpoint{0.890000in}{1.851211in}}%
\pgfpathquadraticcurveto{\pgfqpoint{0.850000in}{1.851211in}}{\pgfqpoint{0.850000in}{1.811211in}}%
\pgfpathlineto{\pgfqpoint{0.850000in}{0.690000in}}%
\pgfpathquadraticcurveto{\pgfqpoint{0.850000in}{0.650000in}}{\pgfqpoint{0.890000in}{0.650000in}}%
\pgfpathlineto{\pgfqpoint{0.890000in}{0.650000in}}%
\pgfpathclose%
\pgfusepath{stroke,fill}%
\end{pgfscope}%
\begin{pgfscope}%
\pgfsetbuttcap%
\pgfsetroundjoin%
\pgfsetlinewidth{1.505625pt}%
\definecolor{currentstroke}{rgb}{0.000000,0.000000,0.000000}%
\pgfsetstrokecolor{currentstroke}%
\pgfsetdash{{5.550000pt}{2.400000pt}}{0.000000pt}%
\pgfpathmoveto{\pgfqpoint{0.930000in}{1.698047in}}%
\pgfpathlineto{\pgfqpoint{1.130000in}{1.698047in}}%
\pgfpathlineto{\pgfqpoint{1.330000in}{1.698047in}}%
\pgfusepath{stroke}%
\end{pgfscope}%
\begin{pgfscope}%
\definecolor{textcolor}{rgb}{0.150000,0.150000,0.150000}%
\pgfsetstrokecolor{textcolor}%
\pgfsetfillcolor{textcolor}%
\pgftext[x=1.490000in,y=1.628047in,left,base]{\color{textcolor}{\sffamily\fontsize{14.400000}{17.280000}\selectfont\catcode`\^=\active\def^{\ifmmode\sp\else\^{}\fi}\catcode`\%=\active\def
\end{pgfscope}%
\begin{pgfscope}%
\pgfsetroundcap%
\pgfsetroundjoin%
\pgfsetlinewidth{1.505625pt}%
\definecolor{currentstroke}{rgb}{0.501961,0.000000,0.501961}%
\pgfsetstrokecolor{currentstroke}%
\pgfsetdash{}{0pt}%
\pgfpathmoveto{\pgfqpoint{0.930000in}{1.412695in}}%
\pgfpathlineto{\pgfqpoint{1.130000in}{1.412695in}}%
\pgfpathlineto{\pgfqpoint{1.330000in}{1.412695in}}%
\pgfusepath{stroke}%
\end{pgfscope}%
\begin{pgfscope}%
\pgfsetbuttcap%
\pgfsetbeveljoin%
\definecolor{currentfill}{rgb}{0.501961,0.000000,0.501961}%
\pgfsetfillcolor{currentfill}%
\pgfsetlinewidth{1.003750pt}%
\definecolor{currentstroke}{rgb}{0.501961,0.000000,0.501961}%
\pgfsetstrokecolor{currentstroke}%
\pgfsetdash{}{0pt}%
\pgfsys@defobject{currentmarker}{\pgfqpoint{-0.039627in}{-0.033709in}}{\pgfqpoint{0.039627in}{0.041667in}}{%
\pgfpathmoveto{\pgfqpoint{0.000000in}{0.041667in}}%
\pgfpathlineto{\pgfqpoint{-0.009355in}{0.012876in}}%
\pgfpathlineto{\pgfqpoint{-0.039627in}{0.012876in}}%
\pgfpathlineto{\pgfqpoint{-0.015136in}{-0.004918in}}%
\pgfpathlineto{\pgfqpoint{-0.024491in}{-0.033709in}}%
\pgfpathlineto{\pgfqpoint{-0.000000in}{-0.015915in}}%
\pgfpathlineto{\pgfqpoint{0.024491in}{-0.033709in}}%
\pgfpathlineto{\pgfqpoint{0.015136in}{-0.004918in}}%
\pgfpathlineto{\pgfqpoint{0.039627in}{0.012876in}}%
\pgfpathlineto{\pgfqpoint{0.009355in}{0.012876in}}%
\pgfpathlineto{\pgfqpoint{0.000000in}{0.041667in}}%
\pgfpathclose%
\pgfusepath{stroke,fill}%
}%
\begin{pgfscope}%
\pgfsys@transformshift{1.130000in}{1.412695in}%
\pgfsys@useobject{currentmarker}{}%
\end{pgfscope}%
\end{pgfscope}%
\begin{pgfscope}%
\definecolor{textcolor}{rgb}{0.150000,0.150000,0.150000}%
\pgfsetstrokecolor{textcolor}%
\pgfsetfillcolor{textcolor}%
\pgftext[x=1.490000in,y=1.342695in,left,base]{\color{textcolor}{\sffamily\fontsize{14.400000}{17.280000}\selectfont\catcode`\^=\active\def^{\ifmmode\sp\else\^{}\fi}\catcode`\%=\active\def
\end{pgfscope}%
\begin{pgfscope}%
\pgfsetroundcap%
\pgfsetroundjoin%
\pgfsetlinewidth{1.505625pt}%
\definecolor{currentstroke}{rgb}{1.000000,0.000000,0.000000}%
\pgfsetstrokecolor{currentstroke}%
\pgfsetdash{}{0pt}%
\pgfpathmoveto{\pgfqpoint{0.930000in}{1.125000in}}%
\pgfpathlineto{\pgfqpoint{1.130000in}{1.125000in}}%
\pgfpathlineto{\pgfqpoint{1.330000in}{1.125000in}}%
\pgfusepath{stroke}%
\end{pgfscope}%
\begin{pgfscope}%
\pgfsetbuttcap%
\pgfsetmiterjoin%
\definecolor{currentfill}{rgb}{1.000000,0.000000,0.000000}%
\pgfsetfillcolor{currentfill}%
\pgfsetlinewidth{1.003750pt}%
\definecolor{currentstroke}{rgb}{1.000000,0.000000,0.000000}%
\pgfsetstrokecolor{currentstroke}%
\pgfsetdash{}{0pt}%
\pgfsys@defobject{currentmarker}{\pgfqpoint{-0.058926in}{-0.058926in}}{\pgfqpoint{0.058926in}{0.058926in}}{%
\pgfpathmoveto{\pgfqpoint{-0.000000in}{-0.058926in}}%
\pgfpathlineto{\pgfqpoint{0.058926in}{0.000000in}}%
\pgfpathlineto{\pgfqpoint{0.000000in}{0.058926in}}%
\pgfpathlineto{\pgfqpoint{-0.058926in}{0.000000in}}%
\pgfpathlineto{\pgfqpoint{-0.000000in}{-0.058926in}}%
\pgfpathclose%
\pgfusepath{stroke,fill}%
}%
\begin{pgfscope}%
\pgfsys@transformshift{1.130000in}{1.125000in}%
\pgfsys@useobject{currentmarker}{}%
\end{pgfscope}%
\end{pgfscope}%
\begin{pgfscope}%
\definecolor{textcolor}{rgb}{0.150000,0.150000,0.150000}%
\pgfsetstrokecolor{textcolor}%
\pgfsetfillcolor{textcolor}%
\pgftext[x=1.490000in,y=1.055000in,left,base]{\color{textcolor}{\sffamily\fontsize{14.400000}{17.280000}\selectfont\catcode`\^=\active\def^{\ifmmode\sp\else\^{}\fi}\catcode`\%=\active\def
\end{pgfscope}%
\begin{pgfscope}%
\pgfsetroundcap%
\pgfsetroundjoin%
\pgfsetlinewidth{1.505625pt}%
\definecolor{currentstroke}{rgb}{1.000000,0.647059,0.000000}%
\pgfsetstrokecolor{currentstroke}%
\pgfsetdash{}{0pt}%
\pgfpathmoveto{\pgfqpoint{0.930000in}{0.839746in}}%
\pgfpathlineto{\pgfqpoint{1.130000in}{0.839746in}}%
\pgfpathlineto{\pgfqpoint{1.330000in}{0.839746in}}%
\pgfusepath{stroke}%
\end{pgfscope}%
\begin{pgfscope}%
\pgfsetbuttcap%
\pgfsetroundjoin%
\definecolor{currentfill}{rgb}{1.000000,0.647059,0.000000}%
\pgfsetfillcolor{currentfill}%
\pgfsetlinewidth{1.003750pt}%
\definecolor{currentstroke}{rgb}{1.000000,0.647059,0.000000}%
\pgfsetstrokecolor{currentstroke}%
\pgfsetdash{}{0pt}%
\pgfsys@defobject{currentmarker}{\pgfqpoint{-0.041667in}{-0.041667in}}{\pgfqpoint{0.041667in}{0.041667in}}{%
\pgfpathmoveto{\pgfqpoint{0.000000in}{-0.041667in}}%
\pgfpathcurveto{\pgfqpoint{0.011050in}{-0.041667in}}{\pgfqpoint{0.021649in}{-0.037276in}}{\pgfqpoint{0.029463in}{-0.029463in}}%
\pgfpathcurveto{\pgfqpoint{0.037276in}{-0.021649in}}{\pgfqpoint{0.041667in}{-0.011050in}}{\pgfqpoint{0.041667in}{0.000000in}}%
\pgfpathcurveto{\pgfqpoint{0.041667in}{0.011050in}}{\pgfqpoint{0.037276in}{0.021649in}}{\pgfqpoint{0.029463in}{0.029463in}}%
\pgfpathcurveto{\pgfqpoint{0.021649in}{0.037276in}}{\pgfqpoint{0.011050in}{0.041667in}}{\pgfqpoint{0.000000in}{0.041667in}}%
\pgfpathcurveto{\pgfqpoint{-0.011050in}{0.041667in}}{\pgfqpoint{-0.021649in}{0.037276in}}{\pgfqpoint{-0.029463in}{0.029463in}}%
\pgfpathcurveto{\pgfqpoint{-0.037276in}{0.021649in}}{\pgfqpoint{-0.041667in}{0.011050in}}{\pgfqpoint{-0.041667in}{0.000000in}}%
\pgfpathcurveto{\pgfqpoint{-0.041667in}{-0.011050in}}{\pgfqpoint{-0.037276in}{-0.021649in}}{\pgfqpoint{-0.029463in}{-0.029463in}}%
\pgfpathcurveto{\pgfqpoint{-0.021649in}{-0.037276in}}{\pgfqpoint{-0.011050in}{-0.041667in}}{\pgfqpoint{0.000000in}{-0.041667in}}%
\pgfpathlineto{\pgfqpoint{0.000000in}{-0.041667in}}%
\pgfpathclose%
\pgfusepath{stroke,fill}%
}%
\begin{pgfscope}%
\pgfsys@transformshift{1.130000in}{0.839746in}%
\pgfsys@useobject{currentmarker}{}%
\end{pgfscope}%
\end{pgfscope}%
\begin{pgfscope}%
\definecolor{textcolor}{rgb}{0.150000,0.150000,0.150000}%
\pgfsetstrokecolor{textcolor}%
\pgfsetfillcolor{textcolor}%
\pgftext[x=1.490000in,y=0.769746in,left,base]{\color{textcolor}{\sffamily\fontsize{14.400000}{17.280000}\selectfont\catcode`\^=\active\def^{\ifmmode\sp\else\^{}\fi}\catcode`\%=\active\def
\end{pgfscope}%
\end{pgfpicture}%
\makeatother%
\endgroup%